\documentclass[10pt,twocolumn,letterpaper]{article}

\usepackage[pagenumbers]{cvpr}

\usepackage{microtype}
\renewcommand{\paragraph}[1]{\vspace{.5em}\noindent\textbf{#1.}}
\usepackage{nicefrac}
\usepackage{multicol, multirow}
\usepackage[table]{xcolor}
\definecolor{1st}{RGB}{102,194,164}
\definecolor{2nd}{RGB}{178,226,226}
\definecolor{3rd}{RGB}{217,248,241}
\definecolor{1stText}{RGB}{57,146,116}

\newcommand{\cmark}{{\color[RGB]{57,146,116}\usefont{U}{pzd}{m}{n}\symbol{51}}}
\newcommand{\xmark}{{\color[RGB]{200,60,50}\usefont{U}{pzd}{m}{n}\symbol{55}}}

\usepackage{dsfont}
\usepackage{enumitem}
\usepackage{bm}

\usepackage{cuted}

\usepackage{tikz}
\tikzset{
  ovlabel/.style={inner sep=1.5pt, font=\scriptsize,
                  text=white, fill=black, fill opacity=0.5, text opacity=1},
  ovpanel/.style={inner sep=0, outer sep=0, anchor=south west},
}
\definecolor{forest}{RGB}{47,93,80}
\definecolor{oxblood}{RGB}{155,45,48}
\definecolor{ocher}{RGB}{201,138,43}
\definecolor{ocherText}{RGB}{150,112,28}
\definecolor{forestTint}{RGB}{224,231,229}

\definecolor{cvprblue}{rgb}{0.21,0.49,0.74}
\usepackage[pagebackref,breaklinks,colorlinks,allcolors=cvprblue]{hyperref}

\title{4EVER: Efficient and Exact Rendering for Volumetric Video with 4D Ellipsoids\vspace{-5.2mm}} %

\author{
Florian Hahlbohm$^{1}$ \hspace{0.5em} Linus Franke$^{2}$ \hspace{0.5em} Vladislav Golyanik$^{3}$ \hspace{0.5em} Martin Eisemann$^{1}$ \hspace{0.5em} Marcus Magnor$^{1, 4}$\\
\small
$^1$TU Braunschweig \quad $^2$University of Würzburg \quad $^3$Max Planck Institute for Informatics, SIC \quad $^4$University of New Mexico\\
\small\textbf{\url{https://fhahlbohm.github.io/4ever}}
}

\begin{document}
\maketitle
\begin{strip}
\centering
\vspace{-1.5cm}
\includegraphics[width=\linewidth]{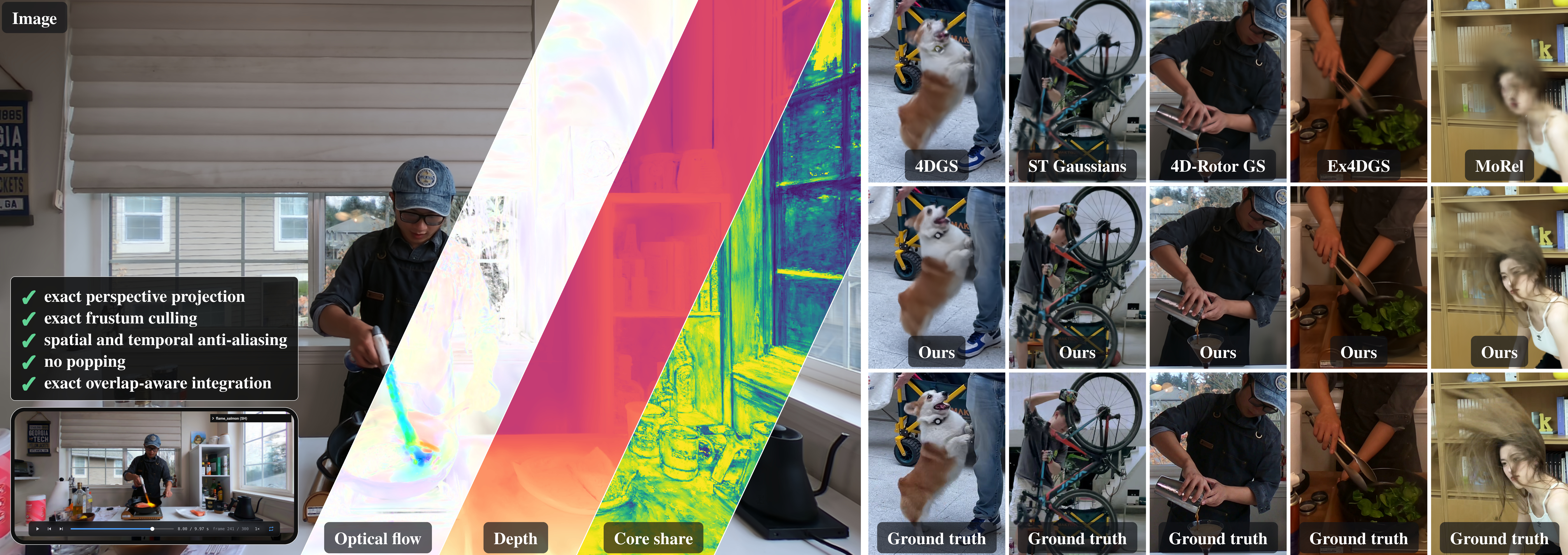}
\vspace{-5mm}
\captionsetup{type=figure}
\captionof{figure}{%
4EVER reconstructs volumetric video with 4D ellipsoids and renders it in real time. For efficient training, we use a core/tail approximation inspired by hybrid transparency~\cite{maule2013hybrid, hahlbohm2025htgs}, where only the core accounts for overlapping primitives, while our WebGL viewer (inset) is fully exact. We show a novel view with optical flow, depth, and core share, \ie, the opacity resolved by the core (left). 4EVER improves novel views of moving content on established benchmarks over baselines~\cite{yang20244dgaussians, li2024spacetime, duan2024rotorgs, lee2024ex4dgs, kwak2026morel} (right).
}\label{fig:teaser}
\end{strip}

\begin{abstract}
Dynamic scene reconstruction methods based on 4D Gaussians suffer from popping, aliasing, and flickering because they inherit the rasterizer approximations of 3D Gaussian Splatting (3DGS).
They project primitives instead of integrating them along the ray, blend them in a per-view global order, and sample them without regard to the pixel footprint or the frame rate.
Exact Volumetric Ellipsoid Rendering (EVER) removes the first two approximations for static scenes, but it requires a ray tracer, and its hard-edged primitives represent fire and smoke poorly.

We present 4EVER, which builds on the observation that exact rendering is affordable in a rasterizer if it is limited to where it matters.
First, to render every frame exactly, we represent the scene with 4D ellipsoids, whose temporal slices are 3D ellipsoids, and give them soft Epanechnikov edges and anti-aliasing in space and time.
Second, to render in real time, our rasterizer integrates the nearest primitives of every pixel exactly, including their overlap, and blends the remainder in a per-tile depth order, and we regularize our models to concentrate the opacity in the nearest primitives.
Third, to capture fast motion, we initialize moving content densely with velocities and give every primitive its own optimizer clock, so that short-lived primitives do not fade out between their observations.
On established benchmarks, 4EVER reduces mean squared error over prior art by more than 10\% on average, both on full frames and in moving regions.
Complemented by a controlled comparison with 4D Gaussians and a WebGL viewer for mobile devices, 4EVER makes exact rendering practical for volumetric video.
\end{abstract}

\section{Introduction}
\label{sec:introduction}
Volumetric video, the reconstruction of a dynamic scene from synchronized multi-view recordings, has become practical with primitive-based radiance fields.
Following 3D Gaussian Splatting (3DGS)~\cite{kerbl3Dgaussians}, dynamic scenes are represented by deforming or tracking a canonical set of 3D Gaussians~\cite{wu20244dgs, yang2024deformable3dgs, luiten2023dynamic}, by equipping 3D Gaussians with an explicit motion and lifetime~\cite{li2024spacetime, wang2025freetimegs}, or by lifting the primitives to four dimensions, so that every frame is a temporal slice of a 4D representation~\cite{yang20244dgaussians, duan2024rotorgs, xu2024temporalgaussianhierarchy}.
All of these methods render with the rasterizer of 3DGS and inherit its approximations.
Gaussians are projected with an affine approximation~\cite{zwicker2001ewasplat}, their overlap along a ray is ignored, and their blending order is a per-view sort by center depth, which is inconsistent across views and causes popping~\cite{radl2024stopthepop}.
Most methods also sample the primitives without regard to the pixel footprint or the frame rate, which causes aliasing and flickering when the sampling rate deviates from that of the training data~\cite{yu2024mip}.
We show that a rasterizer can render volumetric video exactly where it matters and approximately elsewhere, in real time (\cref{fig:teaser}).

Constant-density ellipsoids can instead be rendered \emph{exactly} as a volume, overlap included~\cite{mai2024ever}, but Exact Volumetric Ellipsoid Rendering (EVER) relies on a ray tracer leading to hours of optimization time and achieving low frame rates.
Furthermore, its hard-edged primitives represent volumetric effects such as fire and smoke poorly.
Moreover, the benefits of accurate rendering have so far only been analyzed for static scenes~\cite{huang2024erroranalysis, celarek2025does3dgsneedvolrend}, although popping and flickering are especially disturbing in a video.

Our goal is to render volumetric video exactly and in real time on a single consumer GPU, and to measure what exactness changes for dynamic scenes.
To this end, we present 4EVER, a method that reconstructs dynamic scenes with 4D ellipsoids and renders them in real time, exactly where it matters and approximately where it does not.
We observe that exactness matters most for the nearest primitives of every pixel, because training can concentrate most of the opacity of a pixel in them (\cref{fig:teaser}, core share).
4EVER builds on this observation with three elements.
First, to render every frame exactly, we represent the scene with 4D ellipsoids, since the temporal slice of a 4D ellipsoid is a 3D ellipsoid that can be integrated in closed form.
Since hard-edged ellipsoids represent volumetric effects poorly, we equip them with an Epanechnikov density profile, whose line integral remains closed-form for a single primitive and exact to second order where primitives overlap.
We further make anti-aliasing part of the representation by adapting every primitive to the pixel footprint and the frame rate.

Second, to render in real time, our rasterizer builds on hybrid transparency~\cite{maule2013hybrid, hahlbohm2025htgs}.
For every pixel, the $K$ nearest primitives form a \emph{core} that is integrated exactly, including the overlap between ellipsoids, while the remaining primitives form a \emph{tail} that is blended in a per-tile depth order.
The approximation is thereby confined to the tail, and dedicated training measures concentrate the opacity in the core.

Unlike hybrid transparency for Gaussians~\cite{hahlbohm2025htgs}, our core resolves the overlap of primitives, and our tail is sorted rather than averaged, so it stays accurate even where the core holds little of the opacity.

Third, to capture fast motion, we initialize moving content densely with full 4D shapes and velocities and adapt confidence-aware densification~\cite{hofer2026confidence} to 4D.
Since a primitive that exists for only a few frames is rarely observed during training, the regularizers fade it out between its observations.
We therefore give every primitive its own \emph{observation clock} in Adam~\cite{kingma2014adam}, which updates it only when it is observed, with a larger step the less often this happens.

We evaluate 4EVER on established benchmarks~\cite{li2022neural3dv, wang2025freetimegs} against recent baselines, all retrained under a common protocol that separates reconstruction quality from the photometric calibration of the held-out camera.
4EVER achieves the highest PSNR on both datasets, on full frames and in moving regions.
Furthermore, a controlled comparison of both primitive types across render modes isolates the effect of each approximation and shows the benefits of accurate rendering on dynamic scenes.
Our contributions are: 
\begin{itemize}
\item A 4D ellipsoid representation whose temporal slices are 3D ellipsoids with constant or Epanechnikov density, anti-aliased by pixel footprint and frame rate (\cref{sec:primitives}).
\item A real-time hybrid transparency rasterizer that integrates the nearest primitives of every pixel exactly, including their overlap, with training-time measures that concentrate the opacity in the core (\cref{sec:rendering}).
\item A training scheme for 4D primitives with dense 4D shape initialization, confidence-aware densification adapted to 4D, and an observation clock for the optimizer (\cref{sec:densification}).
\item A controlled analysis of exact versus approximate rendering on dynamic and static scenes under a protocol separating reconstruction from photometric calibration (\cref{sec:exact_vs_approximate}).
\end{itemize}
Our implementation is available on our project page and includes both primitive types, all render modes, and a WebGL viewer based on stochastic transparency~\cite{enderton2010stochastic, kv2025stochasticsplats, windisch2026steadysplats} that plays our reconstructions without the tail approximation, even on mobile devices.

\section{Related Work}
\label{sec:related_work}

\paragraph{Primitive-Based Radiance Fields for Static Scenes}
Neural radiance fields~\cite{mildenhall2020nerf, barron2022mipnerf360} established differentiable volume rendering, and 3DGS~\cite{kerbl3Dgaussians} replaced ray marching with the splatting of anisotropic Gaussians~\cite{zwicker2001ewasplat} in a tile-based rasterizer.
3DGS has since been accelerated~\cite{taming3dgs, hahlbohm2026fastergs} and extended with alternative densification~\cite{kheradmand243dgsmcmc, rotabulo2025revising, hofer2026confidence}, compact appearance~\cite{miazga2026nasgabor, hahlbohm2026appearance}, and other primitives~\cite{huang20242d, held2025trianglesplatting, held2025convexsplatting, von2025linprim, mai2026radmesh, govindarajan2025radfoam, sun2025svraster, liu2025betasplatting}.
These methods target static scenes, and most of them keep the rendering approximations of splatting, both of which we lift for volumetric video.
Our observation clock generalizes visibility-masked Adam updates~\cite{taming3dgs, ding2026adamwgs} from primitives outside the current view to primitives unobserved for most of a sequence.

\begin{figure*}[t]
\centering
\includegraphics[width=\linewidth]{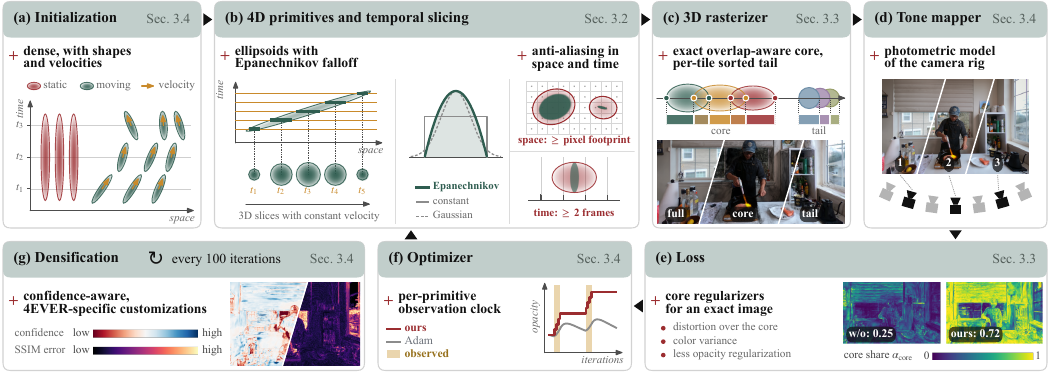}
\caption{%
\textbf{Overview of 4EVER.} (a) We initialize 4D ellipsoids with shapes and velocities. (b) Their slices are 3D ellipsoids with an Epanechnikov falloff, anti-aliased in space and time. (c) We integrate the $K$ nearest primitives per pixel exactly and blend the rest in sorted order. (d) A photometric model per camera~\cite{deutsch2026ppisp} maps radiance to colors. (e) Regularizers raise the opacity $\alpha_\mathrm{core}$ resolved by the core. (f) An observation clock updates primitives only when observed. (g) Densification~\cite{hofer2026confidence} adds primitives at low confidence and high structural error.
}\label{fig:pipeline} 
\end{figure*} 

\paragraph{Accurate Rendering of Primitives}
The affine projection and the global per-view sort of 3DGS cause distortion and popping~\cite{radl2024stopthepop, huang2024erroranalysis}.
Evaluating Gaussians along rays~\cite{moenne2024gaussiantracer, condor2025dontsplat} removes the projection error and has been brought to rasterization for single Gaussians~\cite{wu20253dgut, bulo2025hwraster, huang2026geer, hahlbohm2025htgs, talegaonkar2025vol3dgs, yu2024gof}.
Per-pixel sorting~\cite{radl2024stopthepop, steiner2025aaags} or order-independent transparency~\cite{kv2025stochasticsplats, hou2025sortfree, sun2025stochasticrt} removes the dependence on the global order, and hybrid transparency~\cite{maule2013hybrid, hahlbohm2025htgs} combines both by sorting the nearest fragments of a pixel and averaging the rest.
Closest to our renderer, EVER~\cite{mai2024ever} additionally integrates the overlap of constant-density ellipsoids along the ray, which avoids distortion and popping but requires a hardware ray tracer.
Other works integrate Gaussian or Epanechnikov kernels per primitive~\cite{condor2025dontsplat, blanc2025raygauss, celarek2025does3dgsneedvolrend} or approximate the overlap with moment-based transmittance~\cite{mueller2025moment}, and Softmax-GS~\cite{ziwen2026softmaxgs} re-weights overlapping splats for sharper boundaries.
Unlike EVER, we resolve the overlap of only the $K$ nearest primitives, in a rasterizer, and soften the ellipsoids with an Epanechnikov profile.
In contrast to hybrid transparency for Gaussians~\cite{hahlbohm2025htgs}, our core integrates overlap and our tail is sorted per tile rather than averaged.
As all these works consider static scenes, we extend the question of Celarek~\etal~\cite{celarek2025does3dgsneedvolrend}, who found small gains from volume rendering at a high cost, to dynamic scenes, where ray tracing has so far only simulated camera effects~\cite{liu2025fourdgrt}.

\paragraph{Anti-Aliasing}
Mip-Splatting~\cite{yu2024mip} bounds Gaussians by the pixel footprint at the closest training camera, and analytic~\cite{liang2024analytic, steiner2025aaags} or multi-scale~\cite{yan2024multiscale} integration refines the screen-space filter.
Anti-aliasing of dynamic scenes has received less attention.
Alias-free 4DGS~\cite{chen2025aliasfree4dgs} and Dhiman~\etal~\cite{dhiman2026motionaware} filter 4D Gaussians with a scale-adaptive and a motion-aware filter, respectively, to handle changes in spatial resolution, and RetimeGS~\cite{wang2026retimegs} removes ghosting between frames with flow-supervised trajectories.
These filters act on the image formation of Gaussians, whereas we filter the temporal slice of every primitive with the pixel footprint and bound its temporal extent by the frame spacing.

\paragraph{Dynamic Scene Reconstruction from Video}
Dynamic radiance fields deform a canonical representation~\cite{pumarola2021dnerf, park2021nerfies}, condition the field on time~\cite{li2022neural3dv}, or factorize space and time~\cite{cao2023hexplane, fridovichkeil2023kplanes}.
With Gaussians, deformation fields~\cite{wu20244dgs, yang2024deformable3dgs, xu2024grid4d, wu2025swift4d, kwak2026morel} and per-frame tracking~\cite{luiten2023dynamic} were followed by 3D Gaussians with an explicit motion and lifetime, where the motion is polynomial~\cite{li2024spacetime}, linear~\cite{wang2025freetimegs}, rigid~\cite{yeom2026trigs}, or spline-based~\cite{wang2026retimegs}.
Native 4D Gaussians are instead conditioned on time~\cite{yang20244dgaussians, duan2024rotorgs, gao20257dgs} and have compact~\cite{zhang2025mega, yuan20254dgs1k} and hierarchical~\cite{xu2024temporalgaussianhierarchy} variants, and several methods separate static from dynamic content~\cite{oh2025hybrid3d4dgs, lee2024ex4dgs}.
All of them render with the rasterizer of 3DGS and inherit the approximations that we address.
SharpTimeGS~\cite{liao2026sharptimegs} replaces the Gaussian temporal falloff with a flat-top lifespan to avoid fading, which the slice of a 4D ellipsoid provides by construction.
Closest to our training scheme, FreeTimeGS++~\cite{lee2026freetimegspp} initializes velocities from dense flow at keyframes and corrects the colors of every training camera with a learned affine transformation.
We instead initialize the full 4D shape of every primitive and model the cameras with PPISP~\cite{deutsch2026ppisp}, which restricts the correction to physically plausible camera effects (see \cref{sec:suppl_ppisp}).

\section{Method}
\label{sec:method}
Given synchronized videos of a dynamic scene from $C$ calibrated cameras, with frame $f$ captured at the normalized time $t_f\in[0,1]$, our goal is to render the scene from a novel camera at any time $t\in[0,1]$ in real time.
We represent the scene with 4D primitives and optimize them with the video frames as supervision.
See \cref{fig:pipeline} for our pipeline.

\subsection{Preliminaries}
\label{sec:preliminaries}
A primitive-based radiance field represents a scene with $N$ primitives of density $\rho_i(\mathbf{x})$ and color $\mathbf{c}_i$.
The color of a pixel is the volume rendering integral along its ray $\mathbf{o}+\lambda\mathbf{d}$,
\begin{equation}\label{eq:volume_rendering}
\mathbf{C} = \int_0^\infty \sum_i \rho_i(\lambda)\,\mathbf{c}_i \exp\Bigl(-\int_0^\lambda \sum_j \rho_j(s)\,\mathrm{d}s\Bigr)\,\mathrm{d}\lambda,
\end{equation}
where $\lambda$ is the distance along the ray, $\rho_i(\lambda)=\rho_i(\mathbf{o}+\lambda\mathbf{d})$ the density of primitive $i$ along it, and the exponential the transmittance $T(\lambda)$.
3DGS~\cite{kerbl3Dgaussians} instead projects Gaussians to the image plane with an affine approximation~\cite{zwicker2001ewasplat}, assigns each a per-pixel opacity $\alpha_i$, and alpha-composites them in the order of their center depths,
\begin{equation}\label{eq:alpha_blending}
\mathbf{C} = \textstyle\sum_{i} \alpha_i T_i \mathbf{c}_i, \quad T_i = \prod_{j<i}(1-\alpha_j),
\end{equation}
which ignores the overlap of primitives along the ray and depends on a global sort whose order can change abruptly between views.
EVER~\cite{mai2024ever} uses 3D ellipsoids of constant density $\sigma_i$.
A ray enters and leaves the ellipsoid at $\lambda_i^0$ and $\lambda_i^1$, so the density along the ray is piecewise constant and \cref{eq:volume_rendering} has a closed form.
Between consecutive endpoints of the intersected primitives, the aggregate density $D=\sum_{i\in A}\sigma_i$ of the active set $A$ is constant, and a segment of length $\delta$ contributes $(1-e^{-D\delta})\,T\sum_{i\in A}\sigma_i\mathbf{c}_i/D$, with $T$ at its start.
Without overlap, this reduces to \cref{eq:alpha_blending} with $\alpha_i=1-e^{-\tau}$ for the optical depth ${\tau=\sigma_i(\lambda_i^1-\lambda_i^0)}$ of the chord.

\subsection{4D Ellipsoids}
\label{sec:primitives}
To render every frame exactly, we need a primitive whose temporal slice can be integrated along a ray in closed form.
We therefore propose 4D ellipsoids, which extend the constant-density ellipsoids of EVER~\cite{mai2024ever} to space-time and describe a 3D ellipsoid that moves at constant velocity for a finite lifetime.
They have a mean $\bm{\mu}_i=(\bm{\mu}_i^{x},\mu_i^{t})\in\mathbb{R}^4$, four semi-axes, a 4D rotation given by a left- and a right-isoclinic quaternion~\cite{yang20244dgaussians}, a density, and view-dependent appearance parameters.
Each 4D shape matrix $\bm{\Sigma}_i$ has the spatial, temporal, and mixed blocks $\bm{\Sigma}_i^{xx}$, $\Sigma_i^{tt}$, and $\bm{\Sigma}_i^{xt}$.
To render the frame at time $t$, we slice each ellipsoid at $t$, which yields a 3D ellipsoid with center $\bm{\mu}_{i|t}$ and shape matrix $k_i(t)\,\bm{\Sigma}_{i|t}$,
\begin{align}
\bm{\mu}_{i|t} &= \bm{\mu}_i^{x} + \bm{\Sigma}_i^{xt}\,(\Sigma_i^{tt})^{-1}\,(t-\mu_i^{t}), \label{eq:conditional_mean}\\
\bm{\Sigma}_{i|t} &= \bm{\Sigma}_i^{xx} - \bm{\Sigma}_i^{xt}\,(\Sigma_i^{tt})^{-1}\,\bm{\Sigma}_i^{tx}, \label{eq:conditional_shape}
\end{align}
with $k_i(t) = 1 - (t-\mu_i^{t})^2/\Sigma_i^{tt}$ (see \cref{sec:suppl_derivations}).
The slice moves with the velocity $\bm{\Sigma}_i^{xt}/\Sigma_i^{tt}$ and keeps the shape $\bm{\Sigma}_{i|t}$ up to the scale $k_i(t)$.
It shrinks toward the ends of its lifetime of $2\sqrt{\Sigma_i^{tt}}$ and disappears instead of fading, the geometric analog of the flat-top lifespan of SharpTimeGS~\cite{liao2026sharptimegs}.
A 4D Gaussian shares this parametrization but fades, since conditioning it on $t$ gives the 3D Gaussian with mean $\bm{\mu}_{i|t}$ and covariance $\bm{\Sigma}_{i|t}$ whose opacity is scaled by the temporal marginal $\exp(-\frac{1}{2}(t-\mu_i^t)^2/\Sigma_i^{tt})$~\cite{yang20244dgaussians}.
A 4D ellipsoid has exactly the degrees of freedom of a 3D ellipsoid with a velocity and a lifetime, but its shape, lifetime, and velocity are optimized jointly as semi-axes and a rotation, which we found more stable than separate motion parameters.
As the slice of a constant-density 4D ellipsoid has a constant density, the exact rendering of EVER applies to every frame.
We use spherical harmonics for view-dependent appearance~\cite{kerbl3Dgaussians}.

\paragraph{Density Parametrization}
A constant density has a hard boundary, so content without sharp boundaries, such as fire, smoke, hair, or the blurred edges of fast motion, has to be assembled from many small primitives.
Following Condor~\etal~\cite{condor2025dontsplat}, we also support the Epanechnikov profile $\rho_i(\mathbf{x}) = \sigma_i\,(1-r^2)$ with the normalized radius $r$ of the ellipsoid, whose slice at time $t$ is again an Epanechnikov ellipsoid with peak density $k_i(t)\,\sigma_i$.
Along a ray, the density is a parabola in $\lambda$, so a single primitive is still rendered exactly, and only the overlap of primitives is approximated (\cref{sec:rendering}).
EVER~\cite{mai2024ever} caps the opacity along the shortest axis of a primitive at 0.99, so a primitive can still become fully opaque along its longest axis and block the gradients to everything behind it.
We instead optimize the optical depth $\tau_i$ of the longest chord in log space and cap its opacity $1-e^{-\tau_i}$ at 0.99, so that the cap holds along every chord.
With the chord length $\ell_i$, this gives ${\sigma_i=\tau_i/\ell_i}$ for the constant profile and ${\sigma_i=\nicefrac{3}{2}\,\tau_i/\ell_i}$ for the Epanechnikov profile, which averages $\nicefrac{2}{3}$ of its peak along the chord.

\paragraph{Spatial Anti-Aliasing}
A primitive smaller than a pixel is either hit or missed by the pixel ray, which causes aliasing and starves thin primitives of gradients.
We therefore impose the pixel footprint as a sampling condition on the temporal slice of every primitive, for both primitive types, rather than as a filter on the Gaussian image formation~\cite{chen2025aliasfree4dgs, dhiman2026motionaware}.
Adapting Gaussian anti-aliasing~\cite{yu2024mip} to ellipsoids, we dilate the conditional shape matrix of every primitive by the pixel footprint at its depth $z$, $\hat{\bm{\Sigma}}_{i|t} = k_i(t)\bm{\Sigma}_{i|t} + (z/f)^2\,\mathbf{I}$ with the focal length $f$, and rescale the density by $\sqrt{\det(k_i(t)\bm{\Sigma}_{i|t})/\det\hat{\bm{\Sigma}}_{i|t}}$, which preserves the rendered mass of the primitive.
During optimization, we additionally bound the semi-axes from below by the footprint at the closest training camera~\cite{yu2024mip, hahlbohm2026fastergs}, which we found essential in 4D, where densification creates many tiny primitives that would otherwise die before they receive a gradient.

\paragraph{Temporal Anti-Aliasing}
In time, the sampling condition is the frame rate instead of the pixel footprint.
A primitive that exists for less than the frame spacing is visible in at most one frame and flickers at novel times.
We therefore bound the temporal extent of every primitive from below by two frames, the Nyquist condition for the frame rate.
With this bound, no primitive can appear and vanish between two frames, so the reconstruction can also be rendered at novel times, \eg, in slow motion, without such flickering.

\subsection{Efficient and Exact Rendering}
\label{sec:rendering}
Our renderer has to integrate overlapping primitives exactly while keeping the speed and the bounded per-pixel memory of a tile-based rasterizer.
We extend existing fast tile-based rasterizers~\cite{hahlbohm2025htgs, hahlbohm2026fastergs} to primitives that are integrated along rays.
In a first step, every primitive is sliced at time $t$ and anti-aliased.
For every covered pixel, we transform the ray into the frame in which the ellipsoid is a unit sphere, given by a Cholesky factorization of its shape matrix, and intersect it with this sphere to obtain the entry and exit distances $\lambda_i^0$ and $\lambda_i^1$.
We call the contribution of a primitive to a pixel a \emph{fragment}.
The fragments of a pixel are composited with a variant of hybrid transparency~\cite{maule2013hybrid, hahlbohm2025htgs} (\cref{fig:pipeline}c).
The $K=16$ nearest fragments form the \emph{core}, which we integrate exactly, and all remaining fragments form the \emph{tail}, which we composite behind the core in sorted order.
Nearest refers to the entry distance $\lambda_i^0$ of an ellipsoid and to the depth of a Gaussian.

\paragraph{Exact Core}
For ellipsoids, the core resolves the overlap of its fragments.
We sort their $2K$ entry and exit points, and the pixel color follows from the segment integrals between consecutive points (\cref{sec:preliminaries}).
For a constant density, this is exact.
For the Epanechnikov profile, the aggregate density of a segment is a quadratic, so its optical depth and transmittance are still exact.
Only the split of the segment color among the active fragments requires $\int_0^1 e^{-\tau(x)}\rho_i(x)\,\mathrm{d}x$ with the segment coordinate $x$ and a cubic optical depth $\tau(x)$, which we expand to second order around its linear part.
The resulting pixel colors deviate from a numerical reference by $10^{-6}$ on average.
The backward pass differentiates these integrals, so the geometry of a primitive receives gradients through its entry and exit points.
We derive the integrals, the expansion, and all gradients in \cref{sec:suppl_derivations}, where we also visualize the error.

\paragraph{Sorted Tail}
The tail is alpha-composited behind the core with \cref{eq:alpha_blending}, in a depth order that is established per tile.
This order replaces the global sort by center depth of 3DGS at a small additional cost (\cref{sec:evaluation}).
By construction, it is closer to exact rendering for large primitives, which cover many tiles at different depths.
The tail only receives the transmittance that the core leaves, so both its approximation and its cost matter less the more opacity the core holds.
We therefore skip the tail in the forward and the backward pass for every pixel whose core leaves a transmittance below $10^{-4}$.

\paragraph{Reference Modes}
Our full model uses the exact core and the sorted tail, which we call \emph{hybrid with sorted tail} (``Ours'' in the tables).
To isolate each approximation in our experiments, the same rasterizer provides reference modes that share all other components.
\emph{Global sort} composites all primitives in the center-depth order of 3DGS with \cref{eq:alpha_blending}, which corresponds to EVER without overlap for ellipsoids.
\emph{Hybrid} (HT) is hybrid transparency as in HTGS~\cite{hahlbohm2025htgs}, which keeps the core but accumulates the tail into a weighted average, which does not require sorting and is free of popping.
For ellipsoids, \emph{HT without overlap} selects and composites the core fragments by their chord midpoints.

\paragraph{Concentrating Opacity in the Core}
A model trained with a sorted tail may represent a surface with many faint primitives that are spread over a long depth range, in which case the core holds only part of the opacity.
We measure this with the \emph{core share} $\alpha_\mathrm{core}=1-T_\mathrm{core}$, where $T_\mathrm{core}$ is the transmittance behind the core, averaged over all pixels.
Under plain training, it stays between 0.23 and 0.69 on dynamic scenes (\cref{tab:render_modes}).
We found the opacity regularizer of state-of-the-art densification approaches~\cite{kheradmand243dgsmcmc, hofer2026confidence} to amplify this issue and therefore lower its weight by a factor of ten.
Three training measures raise the core share at no or small cost, and dense initialization raises it further.
First, a distortion loss~\cite{barron2022mipnerf360} over the segments of the core, evaluated in closed form with weights normalized over the core, pulls the mass of a pixel together.
Second, when densification copies an elongated primitive, whose longest semi-axis exceeds 1.5\texttimes\ its second longest, the copies are placed side by side along that axis with their density preserved~\cite{deng2025improvedgs} instead of being stacked as thin layers of fragments.
Third, we penalize the variance of the colors among the fragments of a ray~\cite{jena2025sparfels} in the last third of training~\cite{windisch2026steadysplats}, which pulls tail fragments toward the core or merges them.
Dense initialization (\cref{sec:densification}) helps because it starts from primitives that already sit on surfaces.

\subsection{Initialization, Densification, and Training}
\label{sec:densification}
\paragraph{Dense 4D Initialization}
As in 4DGS~\cite{yang20244dgaussians}, our initial point cloud is reconstructed from the first frame of the videos with multi-view stereo~\cite{schoenberger2016colmap2}.
It is static, so the optimizer would have to grow all moving content by relocation.
We therefore propose a dense initialization for 4D ellipsoids that covers the moving content and sets the 4D shape of every primitive.
For every camera, we first detect the moving pixels, which deviate from the temporal median image by more than a robust noise estimate and typically cover a few percent of a frame.
At every tenth frame, we triangulate these pixels against the three nearest cameras with learned dense correspondences~\cite{edstedt2026romav2}, as in EDGS~\cite{kotovenko2026edgs} for static scenes.
The velocity of a point follows from warping it to the next keyframe within its camera and triangulating it there again.
Every point becomes a 4D primitive with the keyframe time as temporal mean and semi-axes set from the keyframe interval and its nearest-neighbor distance.
Its velocity is written into the mixed block of the shape matrix, $\bm{\Sigma}^{xt}=\mathbf{v}\,\Sigma^{tt}$, so that the primitive already moves correctly.
Static content keeps the first-frame point cloud, with temporal means spread over the sequence.

\paragraph{Confidence-Aware Densification in 4D}
We train under a fixed budget of primitives with the confidence-aware densification of Hofer~\etal~\cite{hofer2026confidence}.
Every 100 iterations, the model grows by 5\% toward the budget, and every 400 iterations, the 10\% of primitives with the lowest rendered impact are relocated.
New and relocated primitives are copied from sources sampled by their positional gradient, their learned confidence, and the structural error of their pixels.
Copies share the optical depth of their source, which preserves the transmittance of constant-density ellipsoids exactly (see \cref{sec:suppl_derivations}).
They are placed within the spatial extent of their source at its time and inherit its time and velocity, since a temporal displacement would move static content outside the observed window, where it receives no gradient.
We also relocate faint ellipsoids, since unlike faint Gaussians they are never culled by the rasterizer and would occupy core slots without contributing opacity.
Finally, a \emph{near-plane rule} relocates every primitive whose bounding sphere comes close to a training camera.
Floaters in this region are invisible to the training views but can cover a held-out one, which we found to be a frequent issue with multi-view capture rigs.

\paragraph{Observation Clock}
A primitive that is visible in only a few frames receives a gradient in few iterations, yet Adam~\cite{kingma2014adam} updates it every iteration.
Regularizers and decaying moments therefore fade transient content out between its observations.
We go beyond the visibility-masked Adam of static scenes~\cite{taming3dgs, ding2026adamwgs} and give every primitive its own clock.
A primitive is updated only when it receives a non-zero gradient, its bias correction counts its own steps, and its step is scaled by $(1/p)^{\gamma}$ with its observation rate $p$, the share of recent iterations in which it had a non-zero gradient.
We use $\gamma=0.5$ for geometry and base color and $\gamma=0$ for view-dependent appearance, so that rarely observed primitives stay close to view-independent.
As the clock also stops the regularizers from fading unobserved primitives, we relocate primitives that received no update for 200 iterations.
As the updates of a primitive no longer depend on how rarely it is observed, we can scale the batch size, \ie, number of views per training iteration, with the length of a sequence, which makes our method applicable to sequences of any length.

\paragraph{Photometric Model of the Rig}
The cameras of a rig may differ in exposure, white balance, and vignetting.
We model these differences per training camera with PPISP~\cite{deutsch2026ppisp}, with a vignetting and a response curve per camera and an exposure and a color correction per frame, and apply the model to the rendered radiance before the loss.

\paragraph{Training Details}
We optimize all parameters for 30k iterations with a batch size of four for sequences of 300 frames, using the fused Adam of Faster-GS~\cite{hahlbohm2026fastergs} with our observation clock.
The loss is the photometric loss of 3DGS, $0.8\,\mathcal{L}_1 + 0.2\,\mathcal{L}_\text{D-SSIM}$, between the output of the photometric model and the video frame, where every pixel is weighted by the rendered confidence~\cite{hofer2026confidence}.
We add the distortion loss and the color variance loss of \cref{sec:rendering} with weights 0.1 and 0.2, the latter from iteration 20k, and the scale regularizer of 3DGS-MCMC~\cite{kheradmand243dgsmcmc} with weight 0.01 until densification ends.
We densify from iteration 600 to 24.9k.
The semi-axes are optimized in log space and clamped to the spatial and the temporal bound of \cref{sec:primitives} after every step.
Since a step in normalized time spans more frames in a longer sequence, we set the learning rate of the temporal means in frames, to 0.01 frames per step.
Otherwise, we use the same learning rates for all scenes (see \cref{sec:suppl_hyperparameters}).

\section{Evaluation}
\label{sec:evaluation}

\subsection{Setup}
\label{sec:setup}
\begin{table*}[tb]
\caption{%
Comparison on Neural 3D Video~\cite{li2022neural3dv} and SelfCap~\cite{wang2025freetimegs}. \cmark\ and \xmark\ indicate whether a method avoids distortion artifacts (dist.), aliasing (AA), popping (pops), and overlap errors (overl.). Time includes the initialization and is measured on an RTX 4090. 4DGS is capped at 2.5M primitives, as it otherwise exceeds 24\,GB of VRAM. The {\setlength{\fboxsep}{0pt}\colorbox{1st}{\strut\,best\,}} and {\setlength{\fboxsep}{0pt}\colorbox{2nd}{\strut\,second best\,}} results are color coded.
}\label{tab:n3dv}
\centering
\setlength\tabcolsep{4.36pt}
\scriptsize
\begin{tabular}{lcccc@{\hskip 6pt}cccccccccccc}
\toprule
 & & & & & \multicolumn{6}{c}{Neural 3D Video~\cite{li2022neural3dv}} & \multicolumn{6}{c}{SelfCap~\cite{wang2025freetimegs}} \\
\cmidrule(lr){6-11} \cmidrule(lr){12-17}
Method & \smash{\rotatebox{90}{dist.}} & \smash{\rotatebox{90}{AA}} & \smash{\rotatebox{90}{pops}} & \smash{\rotatebox{90}{overl.}} & mPSNR$^\uparrow$ & mLPIPS$^\downarrow$ & PSNR$^\uparrow$ & SSIM$^\uparrow$ & LPIPS$^\downarrow$ & Time$^\downarrow$ & mPSNR$^\uparrow$ & mLPIPS$^\downarrow$ & PSNR$^\uparrow$ & SSIM$^\uparrow$ & LPIPS$^\downarrow$ & Time$^\downarrow$ \\
\midrule
4DGS~\cite{yang20244dgaussians}        & \xmark & \xmark & \xmark & \xmark & 32.06 & 0.075 & 33.12 & 0.944 & 0.193 & 2h23m & \cellcolor{2nd}24.47 & \cellcolor{2nd}0.166 & 26.12 & 0.878 & 0.219 & 2h33m \\
4D-Rotor GS~\cite{duan2024rotorgs}     & \xmark & \xmark & \xmark & \xmark & 30.64 & 0.109 & 32.49 & 0.941 & 0.198 & 2h46m & 24.45 & 0.193 & 26.84 & \cellcolor{1st}0.899 & \cellcolor{1st}0.192 & 5h48m \\
ST Gaussians~\cite{li2024spacetime}    & \xmark & \xmark & \xmark & \xmark & \cellcolor{2nd}32.08 & \cellcolor{2nd}0.067 & \cellcolor{2nd}33.44 & \cellcolor{2nd}0.946 & 0.192 & 2h26m & 23.50 & 0.224 & 26.47 & 0.858 & 0.283 & \cellcolor{2nd}34m \\
Ex4DGS~\cite{lee2024ex4dgs}            & \xmark & \xmark & \xmark & \xmark & 29.94 & 0.128 & 32.39 & 0.940 & 0.192 & \cellcolor{1st}57m & 23.14 & 0.257 & 26.81 & 0.856 & 0.247 & 1h01m \\
MoRel~\cite{kwak2026morel}             & \xmark & \xmark & \xmark & \xmark & 29.59 & 0.119 & 32.76 & \cellcolor{2nd}0.946 & \cellcolor{2nd}0.184 & 1h34m & 22.51 & 0.273 & \cellcolor{2nd}26.94 & 0.883 & 0.225 & 2h23m \\
Ours                                   & \cmark & \cmark & \cmark & \cmark & \cellcolor{1st}32.61 & \cellcolor{1st}0.053 & \cellcolor{1st}33.70 & \cellcolor{1st}0.950 & \cellcolor{1st}0.175 & \cellcolor{2nd}1h16m & \cellcolor{1st}25.22 & \cellcolor{1st}0.152 & \cellcolor{1st}27.77 & \cellcolor{2nd}0.897 & \cellcolor{2nd}0.194 & \cellcolor{1st}30m \\
\bottomrule
\end{tabular}
\end{table*}

We run all compared methods ourselves, which ensures that every method receives the same input images and is evaluated in the same way.
Both affect the metrics and can even change the ranking of methods~\cite{kulhanek2025nerfbaselines}.
In particular, downscaling filters differ in how strongly they attenuate high frequencies, and smoother images are easier to fit, which inflates the metrics.

\paragraph{Datasets}
We use the six scenes of the Neural 3D Video dataset~\cite{li2022neural3dv} with 300 frames at 1352$\times$1014 pixels, half the original resolution, and the first camera held out.
We further use six 60-frame windows of SelfCap~\cite{wang2025freetimegs} at about 1885$\times$1060 pixels (1062$\times$1062 for the two bike windows), again with one camera held out.
The dataset authors created these one-second windows by picking the hardest, \ie, fastest-moving, parts of recordings of one to ten minutes, which is exactly what we seek to evaluate.
Both datasets provide videos, which we convert to images with as little resampling as possible and without lossy compression to preserve their quality (see \cref{sec:suppl_frames}).
For the comparison of render modes on static scenes, we use the nine scenes of Mip-NeRF~360~\cite{barron2022mipnerf360} with the established protocol and the pre-downscaled images provided with the dataset.

\paragraph{Photometric Calibration}
Photometric variation affects multi-view video even more than the single-camera captures of static scenes, as every camera of a rig has its own image signal processing.
We therefore equip all baselines with the same PPISP model~\cite{deutsch2026ppisp} that our method uses (\cref{sec:densification}).
As this model has no parameters for the held-out camera, we fit a PPISP model to the frames of this camera for every method with the scene frozen (see \cref{sec:suppl_ppisp}).
This \emph{calibrated} evaluation measures the quality of the reconstruction and not the ability of a method to explain photometric variation.

\paragraph{Metrics}
We report PSNR, SSIM, and LPIPS$_\text{VGG}$~\cite{zhang2018lpips} over all frames of the held-out camera.
As static content covers most of every frame, we also report mPSNR and mLPIPS~\cite{gao2022dycheck}, \ie, PSNR and LPIPS$_\text{Alex}$ inside masks of the moving objects.
A mask covers an object as a whole even if only a part of it moves, \eg, an entire person while only an arm moves, which matches how humans perceive a dynamic scene (see \cref{sec:suppl_masks}).

\paragraph{Baselines}
We compare with 4DGS~\cite{yang20244dgaussians}, 4D-Rotor~GS~\cite{duan2024rotorgs}, ST Gaussians~\cite{li2024spacetime}, Ex4DGS~\cite{lee2024ex4dgs}, and MoRel~\cite{kwak2026morel}.
4DGS shares our 4D Gaussian parametrization, ST Gaussians is the strongest prior method on Neural 3D Video, Ex4DGS is the fastest, and MoRel a strong recent baseline.
Since our protocol requires retraining, we cannot compare with methods whose code was not released at the time of writing~\cite{wang2025freetimegs, yeom2026trigs, chen2025aliasfree4dgs, liao2026sharptimegs, lee2026freetimegspp}.
Beyond the input, dynamic methods differ in initialization, the number of views per iteration, hyperparameters, densification, parametrization, and rendering.
We keep these as published, since they are part of each method, and therefore report training times that include the initialization.
We further note that multiple baselines~\cite{yang20244dgaussians, duan2024rotorgs, li2024spacetime} use per-scene hyperparameters for improved results, which our method and the other baselines~\cite{lee2024ex4dgs, kwak2026morel} avoid.

\paragraph{Our Models}
Our models use the configuration of \cref{sec:densification}, and our 4D models have a budget of one million primitives.
For static scenes, we use the primitive budgets of Liu~\etal~\cite{liu2025betasplatting}.
\looseness=-1 %
We disable the observation clock for these scenes, as we found no benefit when primitives are observed far more frequently than the short-lived primitives of a dynamic scene.

\subsection{Results}
\label{sec:results}
\Cref{tab:n3dv} compares our full model, \ie, 4D ellipsoids with the Epanechnikov profile, with the baselines.
On Neural 3D Video, it achieves the best result in every quality metric, both on full frames and in the moving regions.
On SelfCap, it leads in PSNR and in the moving regions, while 4D-Rotor~GS is slightly ahead in SSIM and LPIPS.
Its margin over the best baseline in mPSNR is largest on SelfCap, where the motion is fastest.
Including the initialization, our model trains in about an hour and a quarter on Neural 3D Video and in about half an hour on SelfCap, whereas most baselines need more than two hours and only Ex4DGS is faster on Neural 3D Video.
It renders at 252\,FPS on Neural 3D Video (\cref{tab:render_modes}).
\Cref{fig:teaser} (right) shows close-ups of fast-moving content from both datasets, each paired with a different baseline.
The baselines blur such content or lose it entirely, \eg, the head of the jumping dog, the tongs, and the hair.
Our reconstructions preserve it, although the finest structures under the fastest motion, such as single strands of hair, remain softer than in the ground truth.
We show more qualitative results and per-scene metrics in \cref{sec:suppl_qualitative,sec:suppl_per_scene} and demonstrate the improvements regarding rendering artifacts in our supplemental video.

\subsection{Exact Versus Approximate Rendering}
\label{sec:exact_vs_approximate}
\label{sec:core_share}
\begin{table*}[tb]
\caption{%
Render modes of our method for all three primitive types on dynamic and static scenes, with \cmark\ and \xmark\ as in \cref{tab:n3dv}. (\cmark) marks rows without core regularization, whose core resolves less of the opacity. Time, without initialization, and FPS are measured on an RTX 4090.
}\label{tab:render_modes}
\centering
\setlength\tabcolsep{2.61pt}
\renewcommand{\arraystretch}{0.96}
\scriptsize
\begin{tabular}{llcccc@{\hskip 6pt}cccccccc@{\hskip 6pt}cccccc}
\toprule
 & & & & & & \multicolumn{8}{c}{Neural 3D Video~\cite{li2022neural3dv} \textit{(dynamic)}} & \multicolumn{6}{c}{Mip-NeRF~360~\cite{barron2022mipnerf360} \textit{(static)}} \\
\cmidrule(lr){7-14} \cmidrule(lr){15-20}
 & Render mode & \smash{\rotatebox{90}{dist.}} & \smash{\rotatebox{90}{AA}} & \smash{\rotatebox{90}{pops}} & \smash{\rotatebox{90}{overl.}} & PSNR$^\uparrow$ & SSIM$^\uparrow$ & LPIPS$^\downarrow$ & mPSNR$^\uparrow$ & mLPIPS$^\downarrow$ & $\alpha_\mathrm{core}$$^\uparrow$ & Time$^\downarrow$ & FPS$^\uparrow$ & PSNR$^\uparrow$ & SSIM$^\uparrow$ & LPIPS$^\downarrow$ & $\alpha_\mathrm{core}$$^\uparrow$ & Time$^\downarrow$ & FPS$^\uparrow$ \\
\midrule
\multirow{4}{*}{\smash{\rotatebox[origin=c]{90}{Gaussian}}} & Global sort w/o AA & \cmark & \xmark & \xmark & \xmark & 33.05 & 0.947 & 0.178 & 32.68 & 0.053 & -- & 32m & 754 & 28.57 & 0.835 & 0.208 & -- & 10m & 344 \\
                                                            & Global sort       & \cmark & \cmark & \xmark & \xmark & 33.21 & 0.947 & 0.178 & 32.70 & 0.054 & -- & 32m & 775 & 28.65 & 0.837 & 0.207 & -- & 10m & 348 \\
                               & HT                & \cmark & \cmark & \cmark & \xmark & 33.56 & 0.948 & 0.173 & 32.29 & 0.054 & 0.68 & 37m & 403 & 28.19 & 0.830 & 0.211 & 0.62 & 12m & 232 \\
                               & Ours              & \cmark & \cmark & (\cmark) & \xmark & 33.68 & 0.949 & 0.173 & 32.44 & 0.053 & 0.66 & 42m & 296 & 28.35 & 0.833 & 0.209 & 0.54 & 14m & 151 \\
\midrule
\multirow{6}{*}{\smash{\rotatebox[origin=c]{90}{Constant}}} & Global sort w/o AA & \cmark & \xmark & \xmark & \xmark & 33.14 & 0.946 & 0.174 & 32.22 & 0.054 & -- & 31m & 885 & 28.30 & 0.833 & 0.218 & -- & 10m & 385 \\
                                                            & Global sort       & \cmark & \cmark & \xmark & \xmark & 33.16 & 0.947 & 0.174 & 32.22 & 0.055 & -- & 32m & 867 & 28.34 & 0.834 & 0.217 & -- & 10m & 382 \\
                               & HT w/o overlap    & \cmark & \cmark & \cmark & \xmark & 32.93 & 0.944 & 0.180 & 31.10 & 0.070 & 0.39 & 37m & 412 & 26.86 & 0.808 & 0.246 & 0.26 & 13m & 220 \\
                               & HT                & \cmark & \cmark & \cmark & \cmark & 33.01 & 0.944 & 0.180 & 31.05 & 0.069 & 0.44 & 47m & 362 & 26.84 & 0.808 & 0.246 & 0.22 & 15m & 216 \\
                                                           & Ours w/o core reg. & \cmark & \cmark & (\cmark) & (\cmark) & 33.48 & 0.948 & 0.174 & 32.17 & 0.056 & 0.69 & 51m & 286 & 28.21 & 0.833 & 0.217 & 0.44 & 16m & 187 \\
                               & Ours              & \cmark & \cmark & \cmark & \cmark & 33.63 & 0.948 & 0.176 & 32.25 & 0.051 & 0.97 & 55m & 325 & 28.00 & 0.828 & 0.218 & 0.91 & 16m & 261 \\
\midrule
\multirow{6}{*}{\smash{\rotatebox[origin=c]{90}{Epanechnikov}}} & Global sort w/o AA & \cmark & \xmark & \xmark & \xmark & 33.13 & 0.947 & 0.178 & 32.52 & 0.056 & -- & 36m & 599 & 28.49 & 0.839 & 0.217 & -- & 12m & 317 \\
                                                            & Global sort       & \cmark & \cmark & \xmark & \xmark & 33.05 & 0.947 & 0.178 & 32.51 & 0.057 & -- & 37m & 571 & 28.50 & 0.840 & 0.216 & -- & 12m & 315 \\
                               & HT w/o overlap    & \cmark & \cmark & \cmark & \xmark & 33.15 & 0.946 & 0.180 & 31.52 & 0.072 & 0.11 & 43m & 297 & 26.88 & 0.813 & 0.245 & 0.10 & 16m & 171 \\
                               & HT                & \cmark & \cmark & \cmark & \cmark & 33.08 & 0.946 & 0.180 & 31.42 & 0.073 & 0.12 & 56m & 221 & 26.92 & 0.814 & 0.245 & 0.11 & 19m & 138 \\
                                                           & Ours w/o core reg. & \cmark & \cmark & (\cmark) & (\cmark) & 33.56 & 0.949 & 0.176 & 32.48 & 0.057 & 0.23 & 58m & 179 & 28.36 & 0.839 & 0.216 & 0.20 & 19m & 107 \\
                               & Ours              & \cmark & \cmark & \cmark & \cmark & 33.70 & 0.950 & 0.175 & 32.61 & 0.053 & 0.67 & 1h04m & 252 & 28.23 & 0.836 & 0.211 & 0.69 & 18m & 173 \\
\bottomrule
\end{tabular}
\end{table*}

\begin{table}[tb]
\centering
\includegraphics[width=\linewidth]{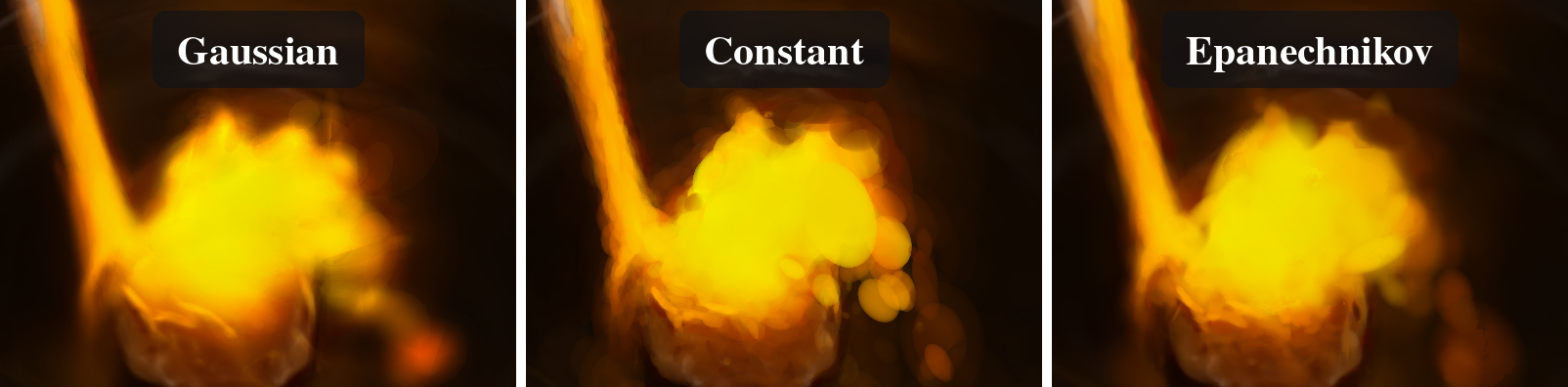}
\captionof{figure}{%
A flame at 4\texttimes\ resolution. Constant-density ellipsoids are clearly visible, an issue that the Epanechnikov profile reduces.
}\label{fig:flame}
\end{table}

\begin{table}[tb]
\caption{%
Ablation of our initialization, observation clock, and the densification method (\cref{sec:densification}) on Neural 3D Video~\cite{li2022neural3dv}.
}\label{tab:ablations}
\centering
\setlength\tabcolsep{1.74pt}
\scriptsize
\begin{tabular}{lccccccc}
\toprule
 & & \multicolumn{3}{c}{Gaussian} & \multicolumn{3}{c}{Epanechnikov} \\
\cmidrule(lr){3-5} \cmidrule(lr){6-8}
Initialization & Clock & mPSNR$^\uparrow$ & mLPIPS$^\downarrow$ & PSNR$^\uparrow$ & mPSNR$^\uparrow$ & mLPIPS$^\downarrow$ & PSNR$^\uparrow$ \\
\midrule
static-only & \xmark & 31.77 & 0.067 & 33.28 & 32.31 & 0.058 & 33.35 \\
static-only & \cmark & 31.43 & 0.071 & 33.18 & 32.03 & 0.063 & 33.47 \\
Ours        & \xmark & 32.01 & 0.059 & 33.52 & 32.63 & 0.052 & 33.76 \\
\midrule
\multicolumn{2}{l}{Ours w/ ADC~\cite{kerbl3Dgaussians}} & 32.19 & 0.053 & 33.15 & 31.93 & 0.063 & 32.95 \\
\multicolumn{2}{l}{Ours w/ MCMC~\cite{kheradmand243dgsmcmc}}                 & 31.99 & 0.062 & 33.68 & 32.28 & 0.057 & 33.58 \\
\midrule
\multicolumn{2}{l}{Ours} & 32.44 & 0.053 & 33.68 & 32.61 & 0.053 & 33.70 \\
\bottomrule
\end{tabular}
\end{table}

\Cref{tab:render_modes} compares the render modes of \cref{sec:rendering} for all three primitive types within our method.
On Neural 3D Video, our mode achieves the highest PSNR for every primitive type, clearly above the global sort, while it is on par in the moving regions for ellipsoids and slightly behind for Gaussians, so its lead stems from the exact ordering of the static content.
HT clearly falls behind in the moving regions for ellipsoids, likely because its order-independent tail fails when the core holds little of the opacity, a dependence that the sorted tail removes.
Our core regularization raises the core share for both ellipsoid profiles, slightly improves PSNR, and accelerates rendering by skipping the tail for more pixels, at slightly increased training time.
The Epanechnikov profile raises mPSNR over the constant density in every mode (\cref{fig:flame}) and matches Gaussians in our mode, at a lower core share and frame rate than the constant profile.
Anti-aliasing has no notable effect on the quality at the training resolution.

On static scenes, in contrast, the global sort achieves the highest PSNR for every primitive type, the core regularization costs a little PSNR, and ellipsoids stay slightly behind Gaussians.
Exactness thus does not pay off in per-frame metrics on static scenes (\cf~\cite{celarek2025does3dgsneedvolrend}), whereas the per-pixel order of our renderer does on dynamic scenes.
Such metrics do not capture popping, however, which our renderer avoids for all content that the core resolves (supplemental video).

\subsection{Ablations}
\label{sec:ablations}
\Cref{tab:ablations} ablates our training scheme for both primitive types and varies the initialization and the observation clock together, since the clock addresses the short-lived primitives that our initialization creates.
Our initialization improves both primitive types on full frames and in the moving regions, with and without the clock.
The clock degrades the moving regions with the static-only initialization, but with our initialization it improves Gaussians in all three metrics and leaves ellipsoids unchanged, while it also enables scaling to sequences of any length (\cref{sec:densification}).
Among the densification methods, confidence-aware densification improves on the adaptive density control (ADC) of 3DGS for both primitive types and surpasses MCMC densification in the moving regions.

\section{Discussion}
\label{sec:discussion}
\vspace{.5em}\noindent\textbf{Does 4DGS Need Exact Volumetric Rendering?}
Per-frame metrics reward accurate rendering on dynamic scenes, where our mode achieves the highest PSNR for every primitive type (\cref{tab:render_modes}), but not on static scenes, in line with Celarek~\etal~\cite{celarek2025does3dgsneedvolrend}.
Such metrics, however, do not capture the popping that accurate rendering removes, which is likely more disturbing in a video, where the camera and the content move, than in a still image.
We therefore consider accurate rendering more important for volumetric video than for static scenes, and our rasterizer makes it affordable during training.

\paragraph{Exact Rendering at Inference}
Our WebGL viewer slices the primitives at the current time, intersects every pixel ray with them, and replaces the core and tail of our rasterizer by stochastic transparency~\cite{enderton2010stochastic, kv2025stochasticsplats}.
In every sample, a fragment survives with a probability equal to its opacity at a depth drawn from its density along the ray, so that the mean over samples converges to \cref{eq:volume_rendering}, including the overlap of primitives.
Alpha correction~\cite{enderton2010stochastic}, stratified survival decisions, and the reprojection of previous frames~\cite{windisch2026steadysplats} reduce the noise enough for interactive frame rates at few samples per frame, at the price of a small bias.
Here, exactness only changes the depth at which a fragment is placed, so it costs little more than approximate blending.
As exact rendering is this cheap at inference, the remaining challenge is to train models for it efficiently, which 4EVER addresses.
We provide additional details in \cref{sec:suppl_viewer}.

\paragraph{Limitations}
During training, the core does not hold all opacity, so part of every pixel is still composited approximately.
This is a limitation of our implementation, as the core size is bounded by the per-pixel memory of our rasterizer, and our viewer resolves all fragments exactly, so only a mismatch between training and viewing remains.
Our rasterizer is also slower than splatting Gaussians in a global order, which we accept since a volumetric video is reconstructed once and played many times, and rendering remains real time.

\section{Conclusion}
\label{sec:conclusion}
We presented 4EVER, which reconstructs volumetric video with 4D ellipsoids and renders it in real time, exactly where it matters.
It achieves the best quality among recent methods, on full frames and in moving regions, while training faster than most of them.
Our controlled comparison shows that accurate rendering pays off on dynamic scenes even in per-frame metrics and avoids artifacts such as popping that these metrics do not capture.
As our rasterizer and viewer make it affordable during training and at inference, we consider accurate rendering a practical default for volumetric video.
\par %
\section*{Acknowledgments}
We thank Mathias Ivanov, Jon Barron, and Thomas Köhler for helpful discussions on individual components of our method.
We also thank Timon Scholz and Kjell Keune for their contributions to the software infrastructure supporting this work.
This work was partially funded by the DFG projects ``Real-Action VR'' (ID 523421583) and ``Increasing Realism of Omnidirectional Videos in Virtual Reality'' (ID 491805996).

{
    \small
    \bibliographystyle{ieeenat_fullname}
    \bibliography{main}
}

\appendix

\section{Derivations}
\label{sec:suppl_derivations}

\subsection{Temporal Slice of a 4D Ellipsoid}
\label{sec:suppl_slice}
\begin{figure}[t]
\centering
\begin{tikzpicture}[x=1cm, y=1cm, >=stealth, font=\footnotesize, line cap=round]
  \def\hl{1.4}  %
  \def\vx{0.9}  %
  \def\sw{1.0}  %
  \def\ft{0.6}  %
  \pgfmathsetmacro{\kt}{1-\ft*\ft}
  \node[anchor=west, inner sep=0] at (-2.45,1.85) {(a)};
  \draw[->] (-1.95,-1.75) -- (-1.95,1.85) node[right, inner sep=3pt] {time};
  \draw[->] (-1.95,-1.75) -- (1.55,-1.75) node[below left, inner sep=2pt] {space};
  \draw[fill=forestTint, draw=forest] plot[domain=0:360, samples=90, smooth cycle] ({\vx*cos(\x)+\sw*sin(\x)}, {\hl*cos(\x)});
  \draw[dashed, black!60] (-\vx,-\hl) -- (\vx,\hl);
  \foreach \f/\c in {0/forest, \ft/oxblood, 0.93/ocher} {
    \pgfmathsetmacro{\hw}{sqrt(1-\f*\f)*\sw}
    \pgfmathsetmacro{\cx}{\f*\vx}
    \pgfmathsetmacro{\cy}{\f*\hl}
    \draw[\c, densely dotted, line width=0.5pt] (-1.95,\cy) -- (\cx-\hw,\cy);
    \draw[\c, line width=1.3pt] (\cx-\hw,\cy) -- (\cx+\hw,\cy);
    \fill[\c] (\cx,\cy) circle (1.4pt);
  }
  \node[left, inner sep=2pt] at (-1.95,0) {$\mu^{t}$};
  \node[left, inner sep=2pt] at (-1.95,\ft*\hl) {$t$};
  \node[below right, inner sep=1.5pt] at (0,0) {$\bm{\mu}$};
  \node[below right, inner sep=1.5pt, oxblood] at (\ft*\vx,\ft*\hl) {$\bm{\mu}_{|t}$};
  \draw[densely dotted, black!50, line width=0.5pt] (\vx,\hl) -- (1.7,\hl);
  \draw[densely dotted, black!50, line width=0.5pt] (-\vx,-\hl) -- (1.7,-\hl);
  \draw[<->, line width=0.4pt] (1.6,-\hl) -- (1.6,\hl) node[midway, right, inner sep=2pt, rotate=90, anchor=north] {lifetime $2\sqrt{\Sigma^{tt}}$};
  \node[anchor=west, inner sep=0] at (2.2,1.85) {(b)};
  \begin{scope}[shift={(4.3,0.94)}]
    \draw[->] (-1.2,0) -- (1.3,0);
    \draw[->] (-1.2,0) -- (-1.2,0.92);
    \draw (-1.25,0.7) -- (-1.15,0.7) node[left, inner sep=3pt] {$\sigma$};
    \foreach \f/\c in {0/forest, \ft/oxblood, 0.93/ocher} {
      \pgfmathsetmacro{\hw}{sqrt(1-\f*\f)*\sw}
      \draw[\c, line width=0.9pt, line join=round] (-\hw,0) -- (-\hw,0.7) -- (\hw,0.7) -- (\hw,0);
    }
    \node at (0.05,0.91) {constant};
    \pgfmathsetmacro{\hwt}{sqrt(\kt)*\sw}
    \draw[oxblood, line width=0.3pt] (-\hwt,0) -- (-\hwt,-0.24) (\hwt,0) -- (\hwt,-0.24);
    \draw[<->, line width=0.4pt, oxblood] (-\hwt,-0.17) -- (\hwt,-0.17);
    \node[below, inner sep=1.5pt, oxblood] at (0,-0.2) {$2\sqrt{k(t)\,\Sigma_{|t}}$};
    \draw[forest, line width=0.3pt] (-\sw,0) -- (-\sw,-0.93) (\sw,0) -- (\sw,-0.93);
    \draw[<->, line width=0.4pt, forest] (-\sw,-0.86) -- (\sw,-0.86);
    \node[below, inner sep=1.5pt, forest] at (0,-0.89) {$2\sqrt{\Sigma_{|t}}$};
  \end{scope}
  \begin{scope}[shift={(4.3,-1.6)}]
    \draw[->] (-1.2,0) -- (1.3,0);
    \draw[->] (-1.2,0) -- (-1.2,0.92);
    \draw (-1.25,0.7) -- (-1.15,0.7) node[left, inner sep=3pt] {$\sigma$};
    \draw[oxblood, densely dotted, line width=0.5pt] (-1.2,0.7*\kt) -- (0,0.7*\kt);
    \draw[oxblood] (-1.25,0.7*\kt) -- (-1.15,0.7*\kt) node[left, inner sep=3pt] {$k(t)\,\sigma$};
    \foreach \f/\c in {0/forest, \ft/oxblood, 0.93/ocher} {
      \pgfmathsetmacro{\k}{1-\f*\f}
      \pgfmathsetmacro{\hw}{sqrt(\k)*\sw}
      \draw[\c, line width=0.9pt] plot[domain=-\hw:\hw, samples=41] (\x, {0.7*\k*(1-(\x/\hw)^2)});
    }
    \node[anchor=south, inner sep=1pt] at (0.05,0.8) {Epanechnikov};
    \node[below, inner sep=3pt] at (0.05,0) {$\mathbf{x}-\bm{\mu}_{|t}$};
  \end{scope}
\end{tikzpicture}
\caption{%
\textbf{Temporal slice of a 4D ellipsoid, shown with one spatial dimension.} (a) The slice at time $t$ is centered at $\bm{\mu}_{|t}$, which moves with the velocity $\mathbf{v}$ along the dashed line, and its extent shrinks with $\sqrt{k(t)}$ until it vanishes at the ends of the lifetime. (b) Density across the three marked slices. A constant density keeps its value $\sigma$, whereas the peak of an Epanechnikov profile decreases to $k(t)\,\sigma$.
}\label{fig:suppl_slice}
\end{figure}
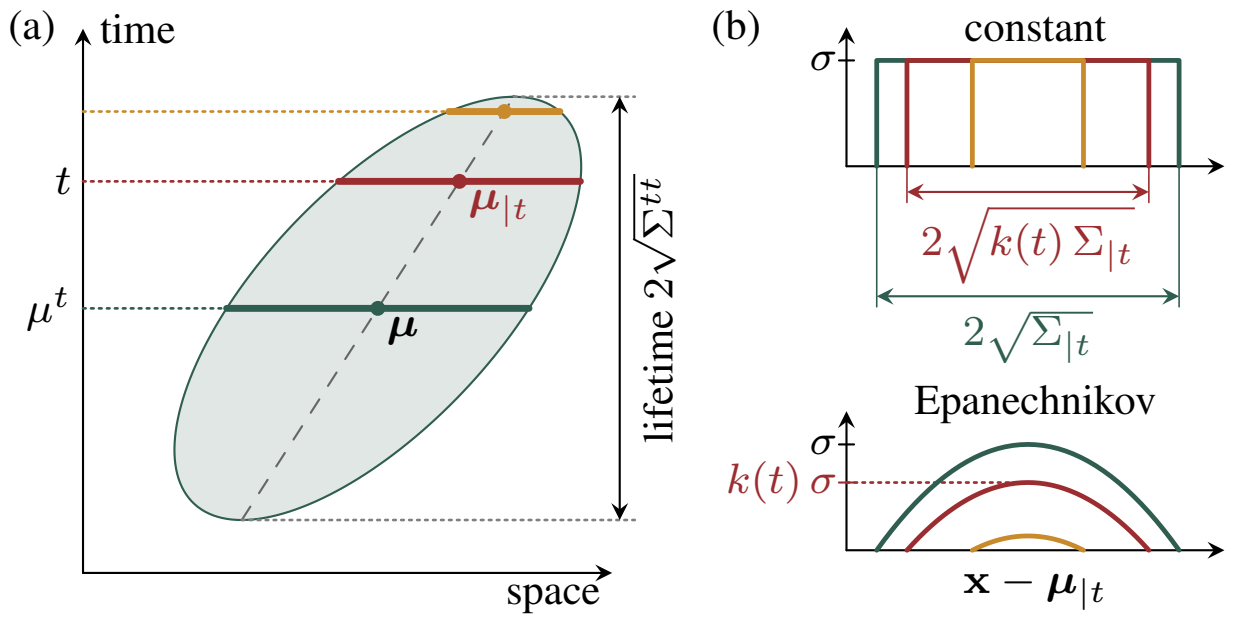

A 4D ellipsoid with mean $\bm{\mu}=(\bm{\mu}^{x},\mu^{t})$ and shape matrix $\bm{\Sigma}$ consists of all points $\mathbf{y}=(\mathbf{x},t)$ of space-time whose squared normalized radius
\begin{equation}\label{eq:suppl_radius}
r^2(\mathbf{y}) = (\mathbf{y}-\bm{\mu})^{\!\top}\,\bm{\Sigma}^{-1}\,(\mathbf{y}-\bm{\mu})
\end{equation}
is at most one.
To slice it at time $t$, we separate space and time in this quadratic form.
Let $\bm{\Sigma}^{xx}$, $\Sigma^{tt}$, and $\bm{\Sigma}^{xt}$ be the spatial, temporal, and mixed blocks of $\bm{\Sigma}$.
With the velocity $\mathbf{v}=\bm{\Sigma}^{xt}/\Sigma^{tt}$ and the matrix
\begin{equation}\label{eq:suppl_conditional_shape}
\bm{\Sigma}_{|t} = \bm{\Sigma}^{xx} - \bm{\Sigma}^{xt}\,(\Sigma^{tt})^{-1}\,\bm{\Sigma}^{tx},
\end{equation}
the shape matrix factors into
\begin{equation}\label{eq:suppl_factorization}
\bm{\Sigma} =
\begin{pmatrix}\mathbf{I} & \mathbf{v}\\ \mathbf{0}^{\!\top} & 1\end{pmatrix}
\begin{pmatrix}\bm{\Sigma}_{|t} & \mathbf{0}\\ \mathbf{0}^{\!\top} & \Sigma^{tt}\end{pmatrix}
\begin{pmatrix}\mathbf{I} & \mathbf{0}\\ \mathbf{v}^{\!\top} & 1\end{pmatrix},
\end{equation}
which is verified by multiplying out the blocks.
The outer factors are shears along time.
The inverse of the left one maps $\mathbf{y}-\bm{\mu}$ to $(\mathbf{x}-\bm{\mu}_{|t},\,t-\mu^{t})$ with the center $\bm{\mu}_{|t}=\bm{\mu}^{x}+\mathbf{v}\,(t-\mu^{t})$, \ie, to coordinates that move with the primitive.
In these coordinates, the quadratic form is block diagonal, and \cref{eq:suppl_radius} becomes
\begin{equation}\label{eq:suppl_separation}
r^2(\mathbf{x},t) = r_{|t}^2(\mathbf{x}) + \frac{(t-\mu^{t})^2}{\Sigma^{tt}},
\end{equation}
with the spatial part $r_{|t}^2(\mathbf{x})=(\mathbf{x}-\bm{\mu}_{|t})^{\!\top}\,\bm{\Sigma}_{|t}^{-1}\,(\mathbf{x}-\bm{\mu}_{|t})$.
The points of the slice thus satisfy $r_{|t}^2(\mathbf{x})\leq k(t)$ with $k(t)=1-(t-\mu^{t})^2/\Sigma^{tt}$.
This is the 3D ellipsoid with center $\bm{\mu}_{|t}$ and shape matrix $k(t)\,\bm{\Sigma}_{|t}$, as stated in \cref{eq:conditional_mean,eq:conditional_shape}.
It shrinks toward the ends of the lifetime and is empty for $k(t)\leq0$ (\cref{fig:suppl_slice}a).

\paragraph{Density of the Slice}
A constant density $\sigma$ inside the 4D ellipsoid is the same constant inside every slice.
For the Epanechnikov profile $\rho=\sigma\,(1-r^2)$, \cref{eq:suppl_separation} gives
\begin{equation}\label{eq:suppl_epanechnikov_slice}
\rho(\mathbf{x},t) = \sigma\,\bigl(k(t)-r_{|t}^2(\mathbf{x})\bigr) = k(t)\,\sigma\,\Bigl(1-\frac{r_{|t}^2(\mathbf{x})}{k(t)}\Bigr),
\end{equation}
where $r_{|t}^2/k(t)$ is the squared normalized radius of the slice.
The slice is therefore again an Epanechnikov ellipsoid, whose peak density $k(t)\,\sigma$ decreases toward the ends of the lifetime while its extent shrinks (\cref{fig:suppl_slice}b).
For a 4D Gaussian $\exp(-\nicefrac{1}{2}\,r^2)$, the same separation yields the 3D Gaussian with covariance $\bm{\Sigma}_{|t}$ and the factor $\exp(-\nicefrac{1}{2}\,(t-\mu^{t})^2/\Sigma^{tt})$, which keeps its extent and only fades~\cite{yang20244dgaussians}.

\paragraph{Stable Evaluation}
\Cref{eq:suppl_conditional_shape} subtracts two nearly equal matrices for a fast primitive, whose spatial extent $\bm{\Sigma}^{xx}$ is dominated by its motion.
In single precision, the difference can turn indefinite, and the affected primitives flicker between training iterations.
We therefore evaluate $\bm{\Sigma}_{|t}$ without subtracting matrices.
Let $\mathbf{m}_1,\dots,\mathbf{m}_4$ be the rows of $\mathbf{M}=\mathbf{R}\mathbf{S}$, where $\mathbf{R}$ is the 4D rotation and $\mathbf{S}$ the diagonal matrix of the semi-axes $s_1,\dots,s_4$, so that $\bm{\Sigma}=\mathbf{M}\mathbf{M}^{\!\top}$.
We remove from every spatial row its component along the temporal row,
\begin{equation}\label{eq:suppl_projection}
\tilde{\mathbf{m}}_j = \mathbf{m}_j - \frac{\mathbf{m}_j^{\!\top}\mathbf{m}_4}{\lVert\mathbf{m}_4\rVert^2}\,\mathbf{m}_4, \quad j\in\{1,2,3\},
\end{equation}
and obtain the entries $\tilde{\mathbf{m}}_j^{\!\top}\tilde{\mathbf{m}}_l$ of $\bm{\Sigma}_{|t}$, a Gram matrix that is positive semi-definite by construction.
As the shears in \cref{eq:suppl_factorization} have unit determinant, $\det\bm{\Sigma}=\det(\bm{\Sigma}_{|t})\,\Sigma^{tt}$, and with $\det\bm{\Sigma}=(s_1s_2s_3s_4)^2$ the determinant of the slice is $\det(k(t)\,\bm{\Sigma}_{|t})=k(t)^3\,(s_1s_2s_3s_4)^2/\Sigma^{tt}$.
It enters the spatial anti-aliasing (\cref{sec:primitives}), where the mass of both density profiles is proportional to $\sigma\sqrt{\det\mathbf{A}}$ for a shape matrix $\mathbf{A}$, so rescaling the density by $\sqrt{\det(k(t)\,\bm{\Sigma}_{|t})/\det\hat{\bm{\Sigma}}_{|t}}$ preserves the mass under the dilation.

\subsection{A Primitive Along a Ray}
\label{sec:suppl_ray}
After slicing and anti-aliasing, a primitive is a 3D ellipsoid with center $\bm{\mu}_{|t}$, shape matrix $\hat{\bm{\Sigma}}_{|t}$, and density $\sigma$, to which the slice and the anti-aliasing contribute the factors of \cref{sec:suppl_slice}.
Let $\mathbf{L}$ be the Cholesky factor of $\hat{\bm{\Sigma}}_{|t}=\mathbf{L}\mathbf{L}^{\!\top}$.
The map $\mathbf{x}\mapsto\mathbf{L}^{-1}(\mathbf{x}-\bm{\mu}_{|t})$ turns the ellipsoid into the unit sphere and the ray into $\tilde{\mathbf{o}}+\lambda\tilde{\mathbf{d}}$ with $\tilde{\mathbf{o}}=\mathbf{L}^{-1}(\mathbf{o}-\bm{\mu}_{|t})$ and $\tilde{\mathbf{d}}=\mathbf{L}^{-1}\mathbf{d}$, while $\lambda$ remains the distance along the ray.
The squared normalized radius along the ray is the parabola
\begin{equation}\label{eq:suppl_ray_radius}
r^2(\lambda) = \lVert\tilde{\mathbf{o}}+\lambda\tilde{\mathbf{d}}\rVert^2 = q + \lVert\tilde{\mathbf{d}}\rVert^2\,(\lambda-\bar{\lambda})^2,
\end{equation}
with the minimum $q=\lVert\tilde{\mathbf{o}}\rVert^2-(\tilde{\mathbf{o}}^{\!\top}\tilde{\mathbf{d}})^2/\lVert\tilde{\mathbf{d}}\rVert^2$ at $\bar{\lambda}=-\tilde{\mathbf{o}}^{\!\top}\tilde{\mathbf{d}}/\lVert\tilde{\mathbf{d}}\rVert^2$.
The ray hits the ellipsoid if $q<1$, and it then enters and leaves at
\begin{equation}\label{eq:suppl_endpoints}
\lambda^{0} = \bar{\lambda}-h, \quad \lambda^{1} = \bar{\lambda}+h, \quad h = \frac{\sqrt{1-q}}{\lVert\tilde{\mathbf{d}}\rVert},
\end{equation}
so $\bar{\lambda}$ is the midpoint and $h$ the half length of its chord.
With the position $w=(\lambda-\bar{\lambda})/h\in[-1,1]$ on the chord, \cref{eq:suppl_ray_radius} reads $r^2=q+(1-q)\,w^2$, and the density along the ray is
\begin{equation}\label{eq:suppl_ray_density}
\rho(\lambda) =
\begin{cases}
\sigma & \text{(constant)},\\
\sigma\,(1-q)\,(1-w^2) & \text{(Epanechnikov)}.
\end{cases}
\end{equation}
The Epanechnikov profile is thus a parabola along every ray, with the peak $\sigma\,(1-q)$ at the midpoint of the chord.
Integrating \cref{eq:suppl_ray_density} over the chord gives the optical depth
\begin{equation}\label{eq:suppl_chord_depth}
\tau = 2\,\sigma h \quad\text{and}\quad \tau = \tfrac{4}{3}\,\sigma\,(1-q)\,h
\end{equation}
of the fragment for the two profiles.
A fragment that is composited without overlap, as in the tail and in the reference modes without overlap, enters \cref{eq:alpha_blending} with $\alpha=1-e^{-\tau}$, which is exact for a single fragment of either profile.

\paragraph{Density Parametrization}
The same integrals define the density of a primitive from the optical depth $\tau$ of a reference chord through its center, where $q=0$.
EVER~\cite{mai2024ever} uses the chord along the shortest semi-axis and caps its opacity $1-e^{-\tau}$ at 0.99 with a scaled sigmoid.
A chord along the longest semi-axis is longer by the ratio of the two semi-axes, and so is its optical depth, so an elongated primitive can still become opaque and block the gradients to everything behind it.
We therefore use the chord along the longest semi-axis, where $2h$ is the chord length $\ell$, and \cref{eq:suppl_chord_depth} gives $\sigma=\tau/\ell$ and $\sigma=\nicefrac{3}{2}\,\tau/\ell$ (\cref{sec:primitives}).
We cap the opacity of this chord at 0.99 as well, and as no chord of a static primitive has a larger optical depth, the cap holds along every ray.
We optimize $\log\tau$ and enforce the cap by clamping this parameter after every optimizer step instead of using a sigmoid.

\subsection{Exact Core}
\label{sec:suppl_core}
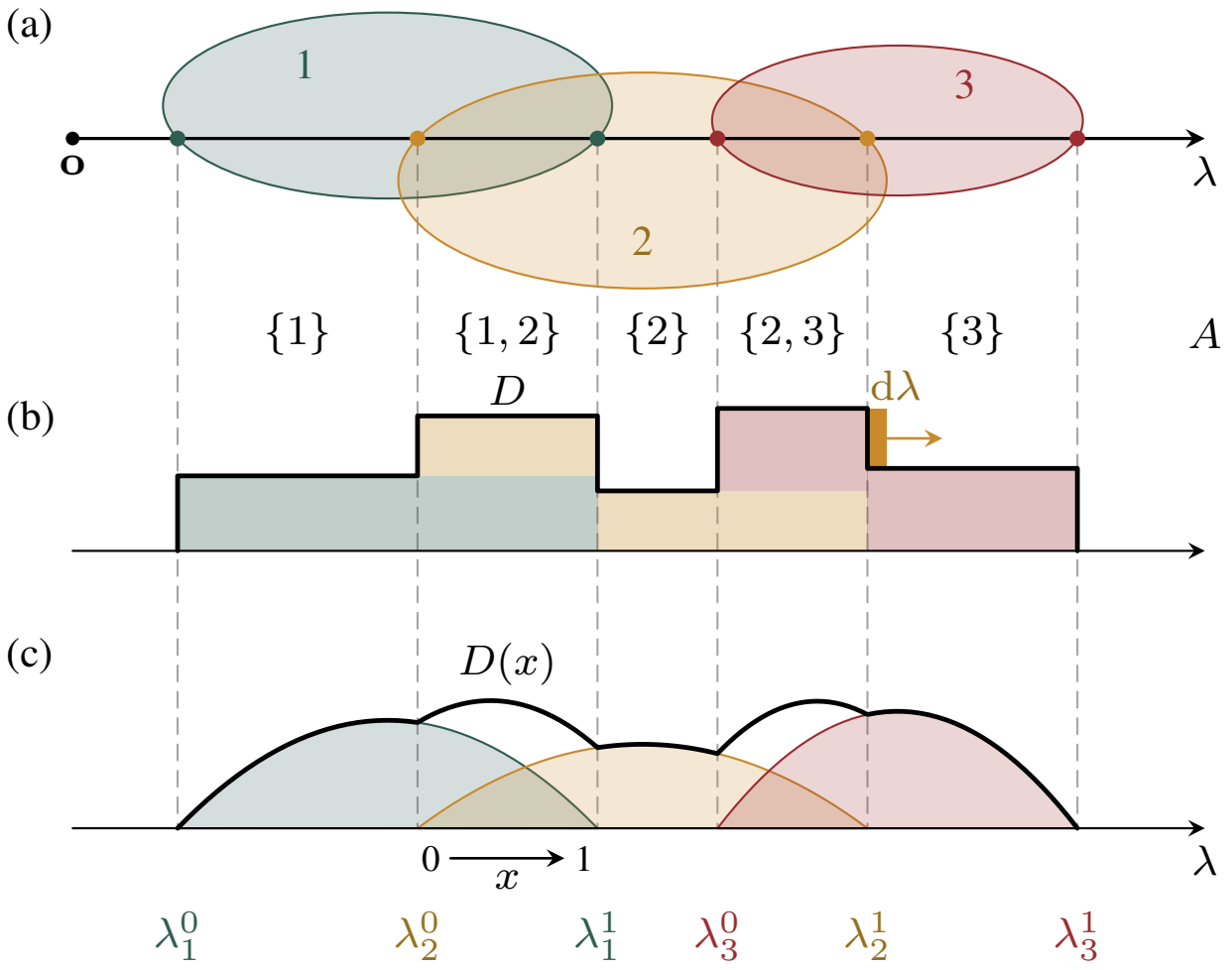
\begin{figure}[t]
\centering
\begin{tikzpicture}[x=1cm, y=1cm, >=stealth, font=\footnotesize, line cap=round, line join=round]
  \def\Aen{0.8} \def\Aex{3.6} %
  \def\Ben{2.4} \def\Bex{5.4} %
  \def\Cen{4.4} \def\Cex{6.8} %
  \pgfmathsetmacro{\Amid}{0.5*(\Aen+\Aex)} \pgfmathsetmacro{\Ahalf}{0.5*(\Aex-\Aen)}
  \pgfmathsetmacro{\Bmid}{0.5*(\Ben+\Bex)} \pgfmathsetmacro{\Bhalf}{0.5*(\Bex-\Ben)}
  \pgfmathsetmacro{\Cmid}{0.5*(\Cen+\Cex)} \pgfmathsetmacro{\Chalf}{0.5*(\Cex-\Cen)}
  \def\yb{-2.75} %
  \def\yc{-4.6}  %
  \foreach \l in {\Aen,\Aex,\Ben,\Bex,\Cen,\Cex} \draw[black!35, densely dashed, line width=0.4pt] (\l,0) -- (\l,\yc);
  \node[anchor=west, inner sep=0] at (-0.35,0.75) {(a)};
  \pgfmathsetmacro{\Arx}{\Ahalf/sqrt(1-(0.22/0.62)^2)}
  \pgfmathsetmacro{\Brx}{\Bhalf/sqrt(1-(0.28/0.72)^2)}
  \pgfmathsetmacro{\Crx}{\Chalf/sqrt(1-(0.12/0.5)^2)}
  \draw[forest, fill=forest, fill opacity=0.2] (\Amid,0.22) ellipse [x radius=\Arx, y radius=0.62];
  \draw[ocher, fill=ocher, fill opacity=0.2] (\Bmid,-0.28) ellipse [x radius=\Brx, y radius=0.72];
  \draw[oxblood, fill=oxblood, fill opacity=0.2] (\Cmid,0.12) ellipse [x radius=\Crx, y radius=0.5];
  \node[forest] at (\Amid-0.55,0.5) {1};
  \node[ocherText] at (\Bmid,-0.68) {2};
  \node[oxblood] at (\Cmid+0.45,0.36) {3};
  \draw[->, line width=0.6pt] (0.1,0) -- (7.65,0) node[below, inner sep=3pt] {$\lambda$};
  \fill (0.1,0) circle (1.3pt) node[below, inner sep=3pt] {$\mathbf{o}$};
  \foreach \l/\c in {\Aen/forest,\Aex/forest,\Ben/ocher,\Bex/ocher,\Cen/oxblood,\Cex/oxblood} \fill[\c] (\l,0) circle (1.5pt);
  \node[inner sep=0] at (7.65,-1.3) {$A$};
  \foreach \l/\r/\s in {\Aen/\Ben/{\{1\}}, \Ben/\Aex/{\{1,2\}}, \Aex/\Cen/{\{2\}}, \Cen/\Bex/{\{2,3\}}, \Bex/\Cex/{\{3\}}} \node[inner sep=0] at ({0.5*(\l+\r)},-1.3) {$\s$};
  \begin{scope}[shift={(0,\yb)}]
    \def\sA{0.5} \def\sB{0.4} \def\sC{0.55}
    \node[anchor=west, inner sep=0] at (-0.35,0.88) {(b)};
    \fill[forest, fill opacity=0.3] (\Aen,0) rectangle (\Aex,\sA);
    \fill[ocher, fill opacity=0.3] (\Ben,\sA) rectangle (\Aex,\sA+\sB);
    \fill[ocher, fill opacity=0.3] (\Aex,0) rectangle (\Bex,\sB);
    \fill[oxblood, fill opacity=0.3] (\Cen,\sB) rectangle (\Bex,\sB+\sC);
    \fill[oxblood, fill opacity=0.3] (\Bex,0) rectangle (\Cex,\sC);
    \fill[ocher] (\Bex,\sC) rectangle (\Bex+0.13,\sC+\sB);
    \draw[->, ocher, line width=0.5pt] (\Bex+0.13,\sC+0.5*\sB) -- (\Bex+0.5,\sC+0.5*\sB);
    \node[ocherText, anchor=south west, inner sep=1.5pt] at (\Bex-0.04,\sC+\sB) {$\mathrm{d}\lambda$};
    \draw[->] (0.1,0) -- (7.65,0);
    \draw[line width=0.9pt] (\Aen,0) -- (\Aen,\sA) -- (\Ben,\sA) -- (\Ben,\sA+\sB) -- (\Aex,\sA+\sB) -- (\Aex,\sB) -- (\Cen,\sB) -- (\Cen,\sB+\sC) -- (\Bex,\sB+\sC) -- (\Bex,\sC) -- (\Cex,\sC) -- (\Cex,0);
    \node[anchor=south, inner sep=2pt] at ({0.5*(\Ben+\Aex)},\sA+\sB) {$D$};
  \end{scope}
  \begin{scope}[shift={(0,\yc)}]
    \def\pA{0.72} \def\pB{0.56} \def\pC{0.78}
    \node[anchor=west, inner sep=0] at (-0.35,1.15) {(c)};
    \draw[forest, fill=forest, fill opacity=0.2] plot[domain=\Aen:\Aex, samples=41] (\x, {\pA*(1-((\x-\Amid)/\Ahalf)^2)}) -- cycle;
    \draw[ocher, fill=ocher, fill opacity=0.2] plot[domain=\Ben:\Bex, samples=41] (\x, {\pB*(1-((\x-\Bmid)/\Bhalf)^2)}) -- cycle;
    \draw[oxblood, fill=oxblood, fill opacity=0.2] plot[domain=\Cen:\Cex, samples=41] (\x, {\pC*(1-((\x-\Cmid)/\Chalf)^2)}) -- cycle;
    \draw[->] (0.1,0) -- (7.65,0) node[below, inner sep=3pt] {$\lambda$};
    \draw[line width=0.9pt]
      plot[domain=\Aen:\Ben, samples=25] (\x, {\pA*(1-((\x-\Amid)/\Ahalf)^2)}) --
      plot[domain=\Ben:\Aex, samples=25] (\x, {\pA*(1-((\x-\Amid)/\Ahalf)^2)+\pB*(1-((\x-\Bmid)/\Bhalf)^2)}) --
      plot[domain=\Aex:\Cen, samples=25] (\x, {\pB*(1-((\x-\Bmid)/\Bhalf)^2)}) --
      plot[domain=\Cen:\Bex, samples=25] (\x, {\pB*(1-((\x-\Bmid)/\Bhalf)^2)+\pC*(1-((\x-\Cmid)/\Chalf)^2)}) --
      plot[domain=\Bex:\Cex, samples=25] (\x, {\pC*(1-((\x-\Cmid)/\Chalf)^2)});
    \node[anchor=south, inner sep=2pt] at ({0.5*(\Ben+\Aex)},0.88) {$D(x)$};
    \draw[->, line width=0.5pt] (\Ben+0.22,-0.2) -- (\Aex-0.22,-0.2) node[midway, below, inner sep=2pt] {$x$};
    \node[inner sep=0] at (\Ben+0.09,-0.2) {\scriptsize 0};
    \node[inner sep=0] at (\Aex-0.09,-0.2) {\scriptsize 1};
    \foreach \l/\c/\s in {\Aen/forest/{\lambda_1^0}, \Ben/ocherText/{\lambda_2^0}, \Aex/forest/{\lambda_1^1}, \Cen/oxblood/{\lambda_3^0}, \Bex/ocherText/{\lambda_2^1}, \Cex/oxblood/{\lambda_3^1}} \node[\c, inner sep=0] at (\l,-0.72) {$\s$};
  \end{scope}
\end{tikzpicture}
\caption{%
\textbf{Fragments along a ray.} (a) The chords of three overlapping primitives. Their sorted endpoints divide the ray into segments with a fixed active set $A$. (b) For constant densities, the aggregate density $D$ is constant on every segment. Moving the exit point of a fragment adds a slab of its density, here for $\lambda_2^1$. (c) For Epanechnikov profiles, $D$ is a quadratic in the coordinate $x$ of every segment.
}\label{fig:suppl_segments}
\end{figure}

The core of a pixel consists of $K$ fragments with densities $\sigma_i$, colors $\mathbf{c}_i$, and chords $[\lambda_i^0,\lambda_i^1]$.
Their $2K$ sorted endpoints divide the ray into segments, and within a segment the active set $A$ of fragments that cover it does not change (\cref{fig:suppl_segments}).
Consider the segment $[\lambda_a,\lambda_b]$ of length $\delta=\lambda_b-\lambda_a$ and let $T_a$ be the transmittance at its start.

\begin{figure*}[t]
\centering
\includegraphics[width=\linewidth]{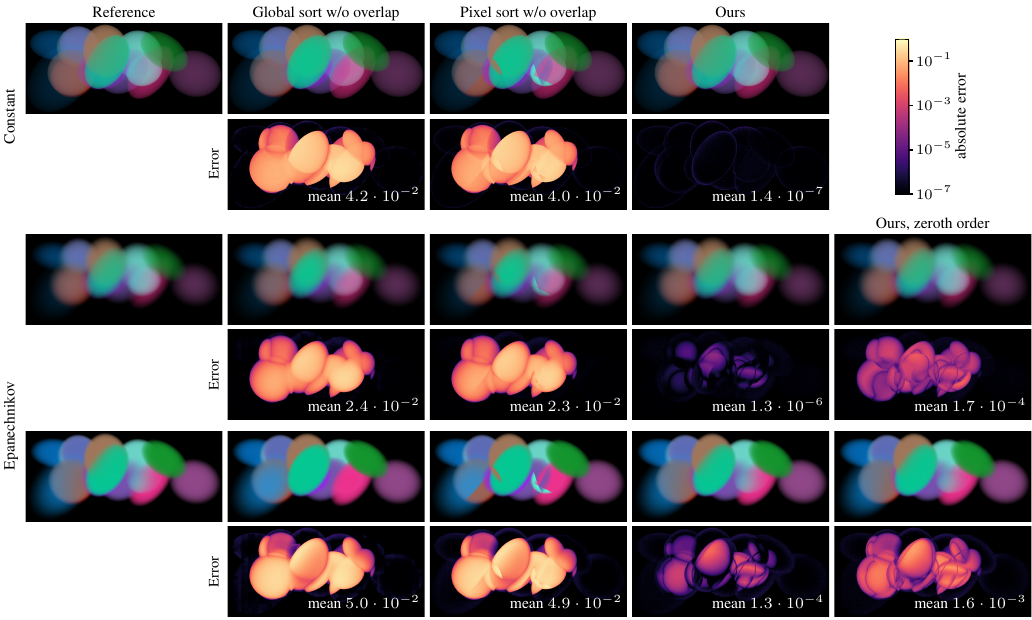}
\caption{%
\textbf{Accuracy of the render modes.} Renderings and per-pixel errors against a float64 quadrature of \cref{eq:volume_rendering} (left) on synthetic scenes of 16 overlapping ellipsoids with constant densities (top), with Epanechnikov profiles (middle), and with more opaque Epanechnikov profiles (bottom). The numbers give the mean over all covered pixels. Global sort and pixel sort without overlap, which is HT without overlap with an empty tail, composite whole fragments and ignore their overlap. Our core needs no expansion for constant densities and is exact up to rounding. For Epanechnikov profiles, its second-order expansion reduces the error of the zeroth order, which keeps only the linear part of the optical depth, by one to two orders of magnitude, and both errors grow with the opacity of the primitives.
}\label{fig:core_error}
\end{figure*}

\paragraph{Constant Densities}
The density on the segment is the constant $D=\sum_{i\in A}\sigma_i$, so the transmittance decays as $T(\lambda)=T_a\,e^{-D(\lambda-\lambda_a)}$, and the segment contributes
\begin{equation}\label{eq:suppl_segment}
\Delta\mathbf{C} = \int_{\lambda_a}^{\lambda_b}\sum_{i\in A}\sigma_i\,\mathbf{c}_i\,T(\lambda)\,\mathrm{d}\lambda = T_a\,\bigl(1-e^{-D\delta}\bigr)\,\bar{\mathbf{c}}
\end{equation}
to \cref{eq:volume_rendering}, with the density-weighted mean color $\bar{\mathbf{c}}=\sum_{i\in A}\sigma_i\mathbf{c}_i/D$ of its active set.
The transmittance $T_b=T_a\,e^{-D\delta}$ remains for the next segment.
A segment thus behaves like a single fragment with the opacity $1-e^{-D\delta}$ and the color $\bar{\mathbf{c}}$, and the core is the alpha blending of \cref{eq:alpha_blending} over its $2K-1$ segments.
The forward pass visits the sorted endpoints once and updates $D$ and $\sum_{i\in A}\sigma_i\mathbf{c}_i$ whenever a fragment enters or leaves.
No approximation is involved, and the rendered core matches a numerical reference up to rounding (\cref{fig:core_error}).
EVER~\cite{mai2024ever} computes the same integral with a ray tracer, whose official implementation does not separate intersections at exactly the same distance along a ray.
This distorts the gradients with respect to the endpoints of coinciding primitives, such as the copies that densification creates, whereas ours remain exact (\cref{sec:suppl_gradients}).

\paragraph{Epanechnikov Profiles}
With the Epanechnikov profile, fragment $i$ has the density $\rho_i(\lambda)=p_i\,(1-w_i^2)$ along the ray, with the peak $p_i=\sigma_i\,(1-q_i)$ and the chord position $w_i=(\lambda-\bar{\lambda}_i)/h_i$ of \cref{sec:suppl_ray}.
The segments and their active sets are the same, but the density now varies within a segment (\cref{fig:suppl_segments}c).
We describe the segment $[\lambda_a,\lambda_b]$ by the coordinate $x\in[0,1]$ with $\lambda=\lambda_a+\delta\,x$.
The chord position of an active fragment is linear in $x$, $w_i=w_i^a+x\,\delta/h_i$ with its value $w_i^a=(\lambda_a-\bar{\lambda}_i)/h_i$ at the start of the segment, so its density $p_i\,(1-w_i^2)$ is the quadratic
\begin{equation}\label{eq:suppl_fragment_quadratic}
\rho_i(x) = \rho_{i,0}+\rho_{i,1}\,x+\rho_{i,2}\,x^2,
\end{equation}
whose coefficients $\rho_{i,n}$ follow by expanding the square.
The aggregate density is then also a quadratic, $D(x)=\sum_{i\in A}\rho_i(x)=D_0+D_1\,x+D_2\,x^2$ with $D_n=\sum_{i\in A}\rho_{i,n}$.

The optical depth between the start of the segment and $x$ is the integral of $\delta\,D$ up to $x$, the cubic
\begin{equation}\label{eq:suppl_cubic}
\tau(x) = \delta\,\bigl(D_0\,x+\tfrac{1}{2}D_1\,x^2+\tfrac{1}{3}D_2\,x^3\bigr).
\end{equation}
The optical depth $\Delta\tau=\tau(1)$ of the whole segment is therefore exact, and so are its opacity $1-e^{-\Delta\tau}$ and the transmittance $T_b=T_a\,e^{-\Delta\tau}$ behind it.

The segment contributes $\Delta\mathbf{C}=T_a\sum_{i\in A}u_i\,\mathbf{c}_i$ to \cref{eq:volume_rendering} with the weights
\begin{equation}\label{eq:suppl_weights}
u_i = \delta\int_0^1 e^{-\tau(x)}\,\rho_i(x)\,\mathrm{d}x,
\end{equation}
which sum to the opacity of the segment.
With the cubic $\tau(x)$ in the exponent, this integral has no closed form.
We split the optical depth into its linear part and a remainder,
\begin{equation}\label{eq:suppl_remainder}
\begin{aligned}
\tau(x) &= \Delta\tau\,x+\varepsilon(x),\\
\varepsilon(x) &= \beta\,(x^2-x)+\gamma\,(x^3-x),
\end{aligned}
\end{equation}
with $\beta=\nicefrac{1}{2}\,\delta D_1$ and $\gamma=\nicefrac{1}{3}\,\delta D_2$.
The remainder vanishes at both ends of the segment and is small if the density varies little across it.
We expand its exponential to second order,
\begin{equation}\label{eq:suppl_expansion}
e^{-\tau(x)} \approx e^{-\Delta\tau\,x}\,\bigl(1-\varepsilon(x)+\tfrac{1}{2}\,\varepsilon(x)^2\bigr).
\end{equation}
The factor in parentheses is a polynomial $\sum_{m=0}^{6}s_m\,x^m$ of degree six with the coefficients $(s_0,\dots,s_6)=(1,\,-a,\,\nicefrac{1}{2}\,a^2-\beta,\,a\beta-\gamma,\,\nicefrac{1}{2}\,\beta^2+a\gamma,\,\beta\gamma,\,\nicefrac{1}{2}\,\gamma^2)$ and $a=-(\beta+\gamma)$.
It is positive for every $\varepsilon$, so no weight can become negative, which a first-order expansion does not guarantee.
Inserting \cref{eq:suppl_fragment_quadratic,eq:suppl_expansion} into \cref{eq:suppl_weights} leaves integrals of a monomial times an exponential,
\begin{equation}\label{eq:suppl_rule}
u_i \approx \tilde{u}_i = \delta\sum_{n=0}^{2}\rho_{i,n}\,F_n, \quad F_n = \sum_{m=0}^{6}s_m\,E_{m+n},
\end{equation}
with the exponential moments $E_n=\int_0^1x^n\,e^{-\Delta\tau\,x}\,\mathrm{d}x$.
The three sums $F_n$ are shared by all fragments of a segment.
Finally, we normalize the weights so that they sum to the exact opacity,
\begin{equation}\label{eq:suppl_epanechnikov_segment}
\Delta\mathbf{C} = T_a\,\bigl(1-e^{-\Delta\tau}\bigr)\,\bar{\mathbf{c}}, \quad \bar{\mathbf{c}} = \frac{\sum_{i\in A}\tilde{u}_i\,\mathbf{c}_i}{\sum_{i\in A}\tilde{u}_i},
\end{equation}
which has the form of \cref{eq:suppl_segment}.
The expansion therefore only affects how the color of a segment is divided among its active fragments, whereas opacity and transmittance are exact.
It is also exact for a segment with a single active fragment and for constant densities, where $\varepsilon=0$.

The zeroth moment is $E_0=(1-e^{-\Delta\tau})/\Delta\tau$, and integration by parts links consecutive moments by $n\,E_{n-1}=\Delta\tau\,E_n+e^{-\Delta\tau}$.
Solved for $E_n$, this recurrence runs upward from $E_0$, but it suffers from cancellation for small $\Delta\tau$.
For $\Delta\tau<8$, we therefore compute the highest moment from the series $E_N=e^{-\Delta\tau}\sum_{m\geq0}\Delta\tau^m\,N!/(N+m+1)!$, whose terms are all positive, and run the recurrence downward.
The forward pass needs the moments up to $E_8$ and the backward pass also $E_9$, since $\mathrm{d}E_n/\mathrm{d}\Delta\tau=-E_{n+1}$.

\Cref{fig:core_error} compares the render modes with a float64 quadrature of \cref{eq:volume_rendering} on synthetic scenes of 16 overlapping ellipsoids, where every fragment is part of the core.
With the second-order expansion, the pixel colors of the Epanechnikov core deviate by $1.3\cdot10^{-6}$ on average, compared to $1.7\cdot10^{-4}$ for the zeroth order, which keeps only the linear part of the optical depth.
Both errors grow with the opacity of the primitives, as $\varepsilon$ scales with their optical depths.
The modes that ignore overlap are off by 2 to $5\cdot10^{-2}$ on the same scenes, and the core for constant densities stays at the rounding error of $10^{-7}$.
The second-order terms add 4\% to the time of a training step in our tests.

\paragraph{Gaussians}
For Gaussians, we evaluate every fragment at its maximal response along the ray, which gives an opacity and a depth, and alpha-composite the core in the order of these depths, ignoring overlap~\cite{hahlbohm2025htgs}.
An exact core is possible with a similar approximation as for the Epanechnikov profile.
The optical depth of a Gaussian along a ray, however, is an error function, which then has to be evaluated for every active fragment at both ends of every segment and also enters the weights of the fragments.
These evaluations are expensive compared to the polynomials of the Epanechnikov profile, so we leave an exact core for Gaussians to future work.

\subsection{Gradients}
\label{sec:suppl_gradients}
The core is an alpha blending of its segments (\cref{sec:suppl_core}), so the gradients of the pixel color $\mathbf{C}$ follow from the chain rule, first with respect to the segments, then to the fragments, and finally to the primitives.

\paragraph{Segments}
We number the segments from front to back.
Segment $s$ has the optical depth $\Delta\tau_s$, the color $\bar{\mathbf{c}}_s$, and the transmittance $T_s$ at its start, so it enters the pixel color with the blending weight $W_s=T_s\,(1-e^{-\Delta\tau_s})$ and leaves the transmittance $T_{s+1}=T_s\,e^{-\Delta\tau_s}$.
Let $\mathbf{C}_{>s}=\mathbf{C}-\sum_{j\leq s}W_j\,\bar{\mathbf{c}}_j$ be the part of the pixel color that originates behind the segment, including the tail and the background.
As this part is proportional to $T_{s+1}$, the derivatives with respect to the color and the optical depth of the segment are
\begin{equation}\label{eq:suppl_segment_gradient}
\frac{\partial\mathbf{C}}{\partial\bar{\mathbf{c}}_s} = W_s, \qquad \frac{\partial\mathbf{C}}{\partial\Delta\tau_s} = T_{s+1}\,\bar{\mathbf{c}}_s-\mathbf{C}_{>s}.
\end{equation}
More optical depth adds the color of the segment at the transmittance behind it and attenuates everything that follows.

\paragraph{Constant Densities}
The optical depth of a segment is $\Delta\tau_s=D_s\,\delta_s$, and its density $D_s$ and color $\bar{\mathbf{c}}_s$ are those of \cref{eq:suppl_segment} for its active set $A_s$.
For a fragment $i$ of this set,
\begin{equation}\label{eq:suppl_fragment_gradient}
\frac{\partial\bar{\mathbf{c}}_s}{\partial\mathbf{c}_i} = \frac{\sigma_i}{D_s}, \quad \frac{\partial\bar{\mathbf{c}}_s}{\partial\sigma_i} = \frac{\mathbf{c}_i-\bar{\mathbf{c}}_s}{D_s}, \quad \frac{\partial\Delta\tau_s}{\partial\sigma_i} = \delta_s.
\end{equation}
Summing over the segments $S_i$ on which the fragment is active gives its blending weight and the gradient of its density,
\begin{align}
\frac{\partial\mathbf{C}}{\partial\mathbf{c}_i} &= \sum_{s\in S_i}W_s\,\frac{\sigma_i}{D_s}, \label{eq:suppl_color_gradient}\\
\frac{\partial\mathbf{C}}{\partial\sigma_i} &= \sum_{s\in S_i}\Bigl(\frac{W_s}{D_s}\,(\mathbf{c}_i-\bar{\mathbf{c}}_s)+\delta_s\,\frac{\partial\mathbf{C}}{\partial\Delta\tau_s}\Bigr). \label{eq:suppl_density_gradient}
\end{align}
The first term of \cref{eq:suppl_density_gradient} changes the color of a segment and the second its opacity.

The endpoints of a fragment change the lengths of the segments.
Moving the exit point $\lambda_i^1$ outward by $\mathrm{d}\lambda$ adds a slab of the fragment (\cref{fig:suppl_segments}b), which to first order is a segment of its own with the optical depth $\sigma_i\,\mathrm{d}\lambda$ and the color $\mathbf{c}_i$, and moving the entry point $\lambda_i^0$ outward removes one.
With \cref{eq:suppl_segment_gradient}, this gives
\begin{equation}\label{eq:suppl_endpoint_gradient}
\begin{aligned}
\frac{\partial\mathbf{C}}{\partial\lambda_i^1} &= \sigma_i\,\bigl(T(\lambda_i^1)\,\mathbf{c}_i-\mathbf{C}_{>\lambda_i^1}\bigr),\\
\frac{\partial\mathbf{C}}{\partial\lambda_i^0} &= -\sigma_i\,\bigl(T(\lambda_i^0)\,\mathbf{c}_i-\mathbf{C}_{>\lambda_i^0}\bigr),
\end{aligned}
\end{equation}
where $T(\lambda)$ is the transmittance at $\lambda$ and $\mathbf{C}_{>\lambda}$ the part of the pixel color that originates behind it.
The same result follows from the lengths of the two segments that share the endpoint, as these segments differ only by the fragment.
\Cref{eq:suppl_endpoint_gradient} also holds if the endpoints of several fragments coincide, as for the copies that densification creates, whereas differentiating the lengths of the sorted segments would attribute a shared endpoint to all of these fragments.
A fragment that is composited as a whole by \cref{eq:alpha_blending} is a segment of its own, so \cref{eq:suppl_segment_gradient} applies with its optical depth $\tau$ and its color $\mathbf{c}_i$.

\paragraph{Epanechnikov Profiles}
The parabola of \cref{eq:suppl_ray_density} vanishes at both ends of the chord, so the endpoints contribute no slab, and the geometry acts through the peak $p_i$, the midpoint $\bar{\lambda}_i$, and the half length $h_i$ instead.
Without overlap, the optical depth $\tau=\nicefrac{4}{3}\,p_i\,h_i$ of \cref{eq:suppl_chord_depth} depends on $p_i$ and $h_i$ only.
In the core, we differentiate the rule of \cref{eq:suppl_rule,eq:suppl_epanechnikov_segment} rather than the exact integral, so that the gradients are consistent with the rendered image.
The rule is a rational function of the coefficients $\rho_{i,n}$ and of the moments $E_n$, whose derivatives are again moments, so all of its derivatives have a closed form.
They are propagated to $p_i$, $\bar{\lambda}_i$, and $h_i$ through \cref{eq:suppl_fragment_quadratic}.

\paragraph{From Fragments to Primitives}
The remaining steps apply the chain rule.
The fragment quantities depend on $\mathbf{L}$ and $\bm{\mu}_{|t}$ through the ray in the frame of the unit sphere (\cref{sec:suppl_ray}), and these depend on the parameters of the 4D ellipsoid through the slice (\cref{sec:suppl_slice}).
The half length $h$ of \cref{eq:suppl_endpoints} has an unbounded derivative for grazing rays, and so has the extent of a slice at the ends of the lifetime.
We therefore skip rays with $q\geq0.99$ and slices with $k(t)\leq0.01$.

\subsection{Splitting a Primitive into Copies}
\label{sec:suppl_split}
Densification replaces a source by $n$ copies, the source itself and $n-1$ new primitives, which share its optical depth $\tau$ (\cref{sec:densification}).
Every copy receives $\tau/n$ and keeps the semi-axes of the source, and therefore its chord length $\ell$.
By \cref{sec:suppl_ray}, the density is proportional to $\tau/\ell$ for both profiles, so every copy has the density $\sigma/n$.
As densities add up in \cref{eq:volume_rendering}, $n$ coinciding copies have the density of the source at every point, and the split changes neither the transmittance nor the color of any ray.
The copies are then displaced within the extent of the source, which redistributes this density.
The opacity of a Gaussian instead scales its response at every pixel, so the same split preserves the transmittance only at the center of the Gaussian, which is why 3DGS-MCMC~\cite{kheradmand243dgsmcmc} also adjusts the scales.

\section{Implementation Details}
\label{sec:suppl_implementation}
\subsection{Rasterizer}
\label{sec:suppl_rasterizer}
Both primitive types and all render modes are implemented in one CUDA extension per primitive type, which shares its structure with Faster-GS~\cite{hahlbohm2026fastergs}: a per-primitive preprocess kernel, instance creation with the exact tile test, a radix sort of the instances for the modes that order them, the blend kernel, and a fused Adam optimizer.
The preprocess kernel is compiled at runtime with NVRTC~\cite{tiny-cuda-nn}, so that the appearance model, the primitive type, and the render mode are compile-time constants, and the fused appearance evaluation of Hahlbohm~\etal~\cite{hahlbohm2026appearance} is inlined into it.
The global sort uses 16\texttimes16 pixel tiles and the per-bucket backward pass of Mallick~\etal~\cite{taming3dgs}. The hybrid modes keep the $K$ core fragments of every pixel in registers and use 8\texttimes8 tiles.
For the sorted tail, the sort key of a primitive in a tile is a 16-bit logarithmic depth between the near and the far plane.
As in HTGS~\cite{hahlbohm2025htgs}, a fragment of a Gaussian enters the core only with an opacity of at least 0.05, and fainter fragments are composited with the tail.
Our ellipsoid modes do not make this approximation.
The core of the hybrid mode with sorted tail is also kept in registers and its per-slot gradient information is written to per-pixel buffers in the forward pass, together with a per-bucket bit mask that marks the core fragments of every pixel. The backward pass replays the instances of a tile from bucket checkpoints like the global sort, takes the gradient of a core fragment from the stored slot instead of recomputing it, and treats the remaining fragments as tail.
Fragments that straddle the near or far plane are clipped exactly to the planes, with the gradient of the clipped endpoint set to zero, rather than dropped.
The near-plane rule and the 3D filter run as fused kernels.

\subsection{Training Details}
\label{sec:suppl_hyperparameters}
For all settings that the main paper does not state, our default configuration uses the values of Hahlbohm~\etal~\cite{hahlbohm2026appearance}, with two exceptions.
The learning rate of the opacity is 0.025, and the weight of the opacity regularizer is 0.001.
The 60-frame windows of SelfCap are trained with one view per iteration.
Static scenes are trained with one view per iteration as well and are initialized from the sparse SfM points of the dataset.
Unless stated otherwise, all settings are shared by static and dynamic scenes, both primitive types, and all render modes.

\paragraph{Initialization}
The dense initialization (\cref{sec:densification}) assigns every point the standard deviations of a Gaussian, and the semi-axes of an ellipsoid are 3.33 times these standard deviations, which is the extent at which a Gaussian falls below the opacity threshold of $\nicefrac{1}{255}$ of 3DGS~\cite{kerbl3Dgaussians}.
It triangulates 1000 of the moving pixels per camera and keyframe.
For a moving point, the temporal standard deviation is half the keyframe interval and the spatial one is half the root mean squared distance to its three nearest neighbors, so its slice has semi-axes of 1.66 times this distance and its lifetime spans 3.33 keyframe intervals.
A static point has a spatial standard deviation of $5\cdot10^{-4}$ times its distance to the nearest training camera~\cite{kotovenko2026edgs}, a temporal mean drawn uniformly from $[-0.1, 1.1]$~\cite{yang20244dgaussians}, and a temporal variance of 0.2~\cite{yang20244dgaussians}.
All primitives start with an opacity of 0.1.
If the initialization exceeds the budget of one million primitives, we keep a random subset of the static points and scale them by the square root of the ratio of the point counts, so that the model starts at the budget and does not grow.
This applies to all scenes of Neural 3D Video, where the initialization holds 1.1 to 1.4 million points and about 50 to 90\% of the static points are kept, whereas the SelfCap windows start with 0.4 to 0.9 million primitives.

\paragraph{Densification Schedule}
The schedule for growing the model and relocating primitives follows Hofer~\etal~\cite{hofer2026confidence}, to whom we refer for details.
Copies of an elongated source are placed side by side along its longest axis~\cite{deng2025improvedgs} (\cref{sec:rendering}), which for a moving source also shifts them in time.
Every copy is 10\% longer than an even division of this axis, so that neighboring copies overlap.

\paragraph{Bounds for Moving Primitives}
The opacity cap and the temporal bound of \cref{sec:primitives} are enforced on the semi-axes of the 4D ellipsoid.
The cap takes twice the longest of the three spatial semi-axes as the chord length $\ell$, and the temporal bound keeps the temporal semi-axis at one frame or more.
For a static primitive, these are the longest chord of every slice and half the lifetime, so both bounds hold as stated.
A moving primitive is tilted in space-time (\cref{fig:suppl_slice}a), so its temporal semi-axis also extends in space and its spatial semi-axes also extend in time.
Its slice can then be longer than $\ell$ along the direction of motion, so that the opacity along this direction can exceed the cap, and its lifetime $2\sqrt{\Sigma^{tt}}$ can be shorter than two frames.
We found neither to be an issue in practice.

\subsection{Inference Rendering}
\label{sec:suppl_viewer}
\paragraph{CUDA Rasterizer}
The inference kernels keep the core of the hybrid modes without overlap in half precision (premultiplied color and opacity as pairs of 16-bit floats, as in HTGS~\cite{hahlbohm2025htgs}), which saves registers and lets the GPU run more threads concurrently.
In all render modes, they only compute the image and omit what the forward pass of training stores for the backward pass (\cref{sec:suppl_rasterizer}), \ie, the bucket checkpoints and the per-pixel gradient information of the core, which avoids the memory traffic of these buffers.

\paragraph{WebGL Viewer}
\begin{table}[t]
\caption{%
\textbf{CUDA rasterizer and WebGL viewer.} FPS and PSNR of our models on flame salmon (Neural 3D Video~\cite{li2022neural3dv}) and flowers (Mip-NeRF~360~\cite{barron2022mipnerf360}). The WebGL viewer renders one stochastic sample per frame on a MacBook Pro and an iPhone~18~Pro.
}\label{tab:viewer}
\centering
\setlength\tabcolsep{5.81pt}
\scriptsize
\begin{tabular}{llcccc}
\toprule
 & & \multicolumn{2}{c}{Flame salmon} & \multicolumn{2}{c}{Flowers} \\
\cmidrule(lr){3-4} \cmidrule(lr){5-6}
 & Renderer & FPS$^\uparrow$ & PSNR$^\uparrow$ & FPS$^\uparrow$ & PSNR$^\uparrow$ \\
\midrule
\multirow{4}{*}{Gaussian} & CUDA, RTX 5090 & 359 & \multirow{2}{*}{30.58} & 236 & \multirow{2}{*}{22.15} \\
 & CUDA, RTX 4090 & 279 & & 179 & \\
 & WebGL, M1 Max & 73 & \multirow{2}{*}{29.16} & 47 & \multirow{2}{*}{21.42} \\
 & WebGL, A20 Pro & 29 & & 22 & \\
\midrule
\multirow{4}{*}{Constant} & CUDA, RTX 5090 & 454 & \multirow{2}{*}{30.06} & 413 & \multirow{2}{*}{21.99} \\
 & CUDA, RTX 4090 & 344 & & 315 & \\
 & WebGL, M1 Max & 112 & \multirow{2}{*}{29.17} & 56 & \multirow{2}{*}{21.28} \\
 & WebGL, A20 Pro & 53 & & 25 & \\
\midrule
\multirow{4}{*}{Epanechnikov} & CUDA, RTX 5090 & 379 & \multirow{2}{*}{30.44} & 317 & \multirow{2}{*}{22.06} \\
 & CUDA, RTX 4090 & 257 & & 219 & \\
 & WebGL, M1 Max & 52 & \multirow{2}{*}{29.64} & 39 & \multirow{2}{*}{21.40} \\
 & WebGL, A20 Pro & 20 & & 17 & \\
\bottomrule
\end{tabular}
\end{table}

\begin{figure*}[t]
\centering
\includegraphics[width=\linewidth]{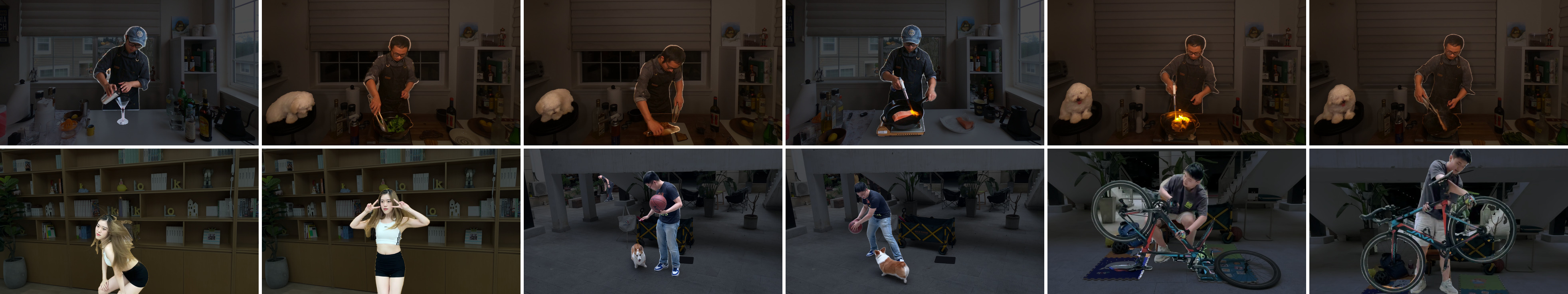}
\caption{%
\textbf{Moving-object masks.} We compute mPSNR and mLPIPS inside masks that cover every moving object as a whole, shown for the held-out cameras of Neural 3D Video~\cite{li2022neural3dv} (top) and SelfCap~\cite{wang2025freetimegs} (bottom) with other content darkened. The masks of Neural 3D Video are obtained automatically from optical flow~\cite{teed2020raft} and SAM~2~\cite{ravi2024sam2}, and those of SelfCap semi-automatically with Cutie~\cite{cheng2024cutie}.
}\label{fig:masks}
\end{figure*}

Our WebGL viewer builds on the web viewer of Hahlbohm~\etal~\cite{hahlbohm2026appearance}, which is based on Three.js, and extends it to the ray-based evaluation of both primitive types in 3D and 4D.
An exported model stores 24 bytes of geometry per 4D primitive in a compact layout similar to Spark~\cite{spark2025}, namely a half-precision spatial mean, a 16-bit temporal mean, 8-bit logarithmic codes for the four semi-axes, two 24-bit quaternions for the 4D rotation, and the 8-bit base color and opacity.
Whenever the time changes, a slice pass evaluates one texel per primitive in a fragment shader and writes the center, the shape matrix, and the density scale of its slice (\cref{sec:suppl_slice}) to floating-point textures.
A second pass of this kind evaluates the view-dependent colors whenever the camera moves.
The render pass draws one silhouette quad per primitive~\cite{gumhold2003ellipsoids}, built from the Cholesky factor of the shape matrix of the slice.
As in \cref{sec:suppl_ray}, the ellipsoid is the unit sphere in this frame, so the fragment shader obtains the chord of its ray and the optical depth of \cref{eq:suppl_chord_depth} from a few dot products, with the anti-aliasing, the grazing cutoff, and the clipping to the near and far plane of the CUDA rasterizer.
As the slices change with every frame of a video, the viewer does not sort 4D primitives and composites them with alpha-corrected stochastic transparency~\cite{enderton2010stochastic} in all render modes, whereas 3D models trained with the global sort are sorted on the CPU and blended back to front.
Per sample, a first pass draws eight stratified survival decisions for every fragment and keeps the nearest surviving depth of each draw.
A second pass accumulates the colors of all fragments, weighted by their opacity and by the share of the eight depths that do not lie in front of them, and a resolve scales the result to the exact coverage of the pixel.
For models trained with overlap, the depths are drawn from the distribution of the first scattering event inside the chord, \ie, with free-flight distance sampling as in StochasticSplats~\cite{kv2025stochasticsplats}, which accounts for the overlap of the primitives.
The image is averaged over the samples while the view is still, and under camera motion and playback, a temporal reprojection as in SteadySplats~\cite{windisch2026steadysplats}, extended by the displacement of the slice centers between frames, continues the average.
The viewer thus renders the per-pixel order without the core and tail of our render mode, which differs only where more than $K$ fragments overlap before a pixel saturates.

In \cref{tab:viewer}, the CUDA rasterizer renders at 179 to 454\,FPS.
With one sample per frame, the WebGL viewer reaches 39 to 112\,FPS on a laptop and 17 to 53\,FPS on a phone.
Its PSNR, which is nearly identical on both devices, is 0.7 to 1.4\,dB below the CUDA rasterizer, and averaging more samples recovers part of this difference.

\section{Evaluation Setup}
\label{sec:suppl_evaluation}
\subsection{Frame Extraction}
\label{sec:suppl_frames}
Neural 3D Video~\cite{li2022neural3dv} and SelfCap~\cite{wang2025freetimegs} provide videos, and every resampling or lossy compression step between a video and the training images removes detail, which makes the images easier to fit and inflates the metrics, while the reconstructions contain less high-frequency detail.
We therefore decode every video once, resample every frame only once with a Lanczos filter, and store it as a lossless PNG file.
For Neural 3D Video, this single step is the downscaling to half resolution.
For SelfCap, the frames also have to be undistorted, so the single step maps every decoded frame directly to the pixel grid of a COLMAP undistortion followed by a downscaling by a factor of two, with no downscaling for the bike windows.
This step resamples the luma and chroma planes of the video separately, so that the chroma, which the videos store at half resolution, is never upsampled first.
All methods, \ie, ours and the baselines, load the extracted frames without further resampling and thus see identical images.
We also found the videos of SelfCap to shake by a few pixels from frame to frame, most strongly in the dance sequences.
Our attempts to remove this shake, namely aligning every frame with a homography or a translation and fitting a camera pose per frame, did not help, so all methods use the uncorrected frames.

\subsection{Photometric Calibration}
\label{sec:suppl_ppisp}
We use PPISP~\cite{deutsch2026ppisp}, which models only the photometric effects of a physical camera and is thus better constrained than an affine color correction per camera~\cite{lee2026freetimegspp}.
It describes a camera with a vignetting, a response curve, and an exposure and a color correction per frame, which a controller predicts from the rendering for novel views.
In a rig, every camera has its own processing, and the cameras differ far more from each other than a camera changes over time.
With one model for all cameras, as in the standard setup for a single camera, the per-frame exposure and color corrections absorb these differences and are almost constant per camera.
The controller cannot infer such an offset from a rendering and outputs a constant for the held-out camera that misses its offset.
We therefore give every training camera its own vignetting and response curve, keep the exposure and the color correction per frame, and train no controller.
In early experiments, we found that these per-camera models also improve the reconstruction itself.

\paragraph{Held-Out Camera}
The held-out camera then has no parameters.
For all results on Neural 3D Video and SelfCap in the main paper, we therefore calibrate it for every method, \ie, we freeze the scene and fit a fresh single-camera model with per-frame exposure and color, one vignetting, and one response curve to its frames, with the photometric loss and regularizers of training for 5k steps.
This calibration separates the quality of the reconstruction from the error of guessing the response of an unseen camera.
Without calibration, we use the per-parameter median over the training cameras, and \cref{tab:per_scene,tab:per_scene_selfcap} compare both for every scene and method.
In early experiments, we found the median to be a better guess than the controller of a single-camera model or the three nearest training cameras.

\subsection{Moving-Object Masks}
\label{sec:suppl_masks}
\begin{table}[t]
\caption{%
\textbf{Core size and core regularization.} Means over the scenes of Neural 3D Video~\cite{li2022neural3dv} for 4D ellipsoids with the Epanechnikov profile in our mode, with the metrics of \cref{tab:render_modes}. The upper block varies the number $K$ of core fragments, and the lower block removes one of the measures of \cref{sec:rendering} at a time.
}\label{tab:core_share_suppl}
\centering
\setlength\tabcolsep{5.36pt}
\scriptsize
\begin{tabular}{lccccc}
\toprule
Configuration & PSNR$^\uparrow$ & SSIM$^\uparrow$ & LPIPS$^\downarrow$ & $\alpha_\mathrm{core}$$^\uparrow$ & Time$^\downarrow$ \\
\midrule
Ours ($K=16$)                  & 33.70 & 0.950 & 0.175 & 0.67 & 1h04m \\
$K=32$                         & 33.75 & 0.950 & 0.175 & 0.93 & 2h33m \\
$K=64$                         & 33.69 & 0.950 & 0.176 & 0.99 & 6h39m \\
\midrule
w/o distortion loss            & 33.77 & 0.950 & 0.175 & 0.59 & 55m \\
w/o long-axis split            & 33.78 & 0.950 & 0.175 & 0.57 & 1h05m \\
w/o color variance loss        & 33.68 & 0.950 & 0.175 & 0.58 & 1h03m \\
opacity regularizer 0.01       & 33.83 & 0.950 & 0.175 & 0.41 & 1h06m \\
w/o core regularization        & 33.56 & 0.949 & 0.176 & 0.23 & 58m \\
\bottomrule
\end{tabular}
\end{table}

\begin{table}[t]
\caption{%
\textbf{Anti-aliasing.} Mean PSNR over the scenes of Neural 3D Video~\cite{li2022neural3dv} for the held-out camera at the training resolution (1\texttimes), at the native resolution of the videos (2\texttimes), and at half the training resolution (0.5\texttimes). We remove the density rescaling, the lower bound of the semi-axes (3D filter), and the lower bound of the temporal extent (temporal floor) of \cref{sec:primitives} one at a time and all at once, which is the configuration w/o AA of \cref{tab:render_modes}.
}\label{tab:antialiasing}
\centering
\setlength\tabcolsep{12.51pt}
\scriptsize
\begin{tabular}{lccc}
\toprule
Configuration & 1\texttimes & 2\texttimes & 0.5\texttimes \\
\midrule
Ours                                    & 33.70 & 33.00 & 33.58 \\
w/o density rescaling                   & 33.84 & 32.19 & 31.54 \\
w/o 3D filter                           & 33.88 & 33.10 & 33.78 \\
w/o temporal floor                      & 33.82 & 33.11 & 33.68 \\
w/o all three (w/o AA in \cref{tab:render_modes}) & 33.80 & 31.43 & 31.34 \\
\bottomrule
\end{tabular}
\end{table}

\begin{figure}[t]
\centering
\includegraphics[width=\linewidth]{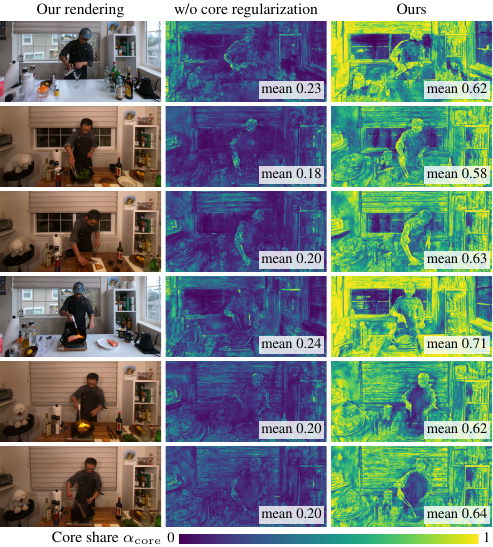}
\caption{%
\textbf{Core share.} Per-pixel core share $\alpha_\mathrm{core}$ and its mean for one test view per scene of Neural 3D Video~\cite{li2022neural3dv}, for the last and the first row of \cref{tab:core_share_suppl}, \ie, without and with core regularization.
}\label{fig:core_share}
\end{figure}

Our evaluation uses dynamic region masks that cover every moving object as a whole (\cref{fig:masks}), \eg, a person together with the objects they hold while only an arm moves.
Thresholding the change of every pixel cannot provide such masks, as it only marks the pixels that change.
For Neural 3D Video, we therefore segment object instances in the videos of the held-out camera and decide per instance whether it moves, without any manual input.
Optical flow~\cite{teed2020raft} between frames five apart seeds SAM~2~\cite{ravi2024sam2} instances at keyframes, and each instance is tracked over the sequence.
An instance is kept if it ever moves in place or changes its silhouette, judged by robust change statistics against a per-object background estimate, and is then masked in every frame.
The final masks cover 6 to 10\% of the frame and are closed and dilated by a few pixels in space and time, so that the boundary band, where reconstruction errors concentrate, lies inside the region.
For SelfCap, the automatic masks were less reliable, so we annotated the masks of the held-out camera semi-automatically with Cutie~\cite{cheng2024cutie}.
Following DyCheck~\cite{gao2022dycheck}, mPSNR is computed over the masked pixels and mLPIPS on the masked image with AlexNet.

\section{Further Experiments and Results}
\label{sec:suppl_experiments}
\subsection{Core Share}
\label{sec:suppl_core_share}
\Cref{tab:core_share_suppl} extends the comparison of \cref{tab:render_modes} to larger cores and to the four individual measures of our core regularization (\cref{sec:rendering}), \ie, the distortion loss, the long-axis split, the color variance loss, and the weight of the opacity regularizer, which we lower from the 0.01 of Hofer~\etal~\cite{hofer2026confidence} to 0.001.
The last row reverts all four measures, which is the configuration without core regularization of \cref{tab:render_modes}.
Compared to it, our default model raises the core share (\cref{fig:core_share}) at slightly higher PSNR and training time.

\subsection{Anti-Aliasing}
\label{sec:suppl_antialiasing}
\Cref{tab:antialiasing} ablates the density rescaling, the 3D filter, and the temporal floor at three resolutions.
All configurations perform similarly at the training resolution.
At the others, the PSNR drops clearly without the density rescaling.
The 3D filter and the temporal floor target spatiotemporal extrapolation from the training views, \eg, close-ups or slow motion, and removing either does not lower the PSNR here.

\begin{figure*}[p]
\centering
\setlength\tabcolsep{0pt}
\footnotesize
\begin{tabular}{*{7}{p{0.14285714\linewidth}<{\centering}}@{}}
4DGS~\cite{yang20244dgaussians} & 4D-Rotor GS~\cite{duan2024rotorgs} & ST Gaussians~\cite{li2024spacetime} & Ex4DGS~\cite{lee2024ex4dgs} & MoRel~\cite{kwak2026morel} & Ours & Ground truth
\end{tabular}
\includegraphics[width=\linewidth]{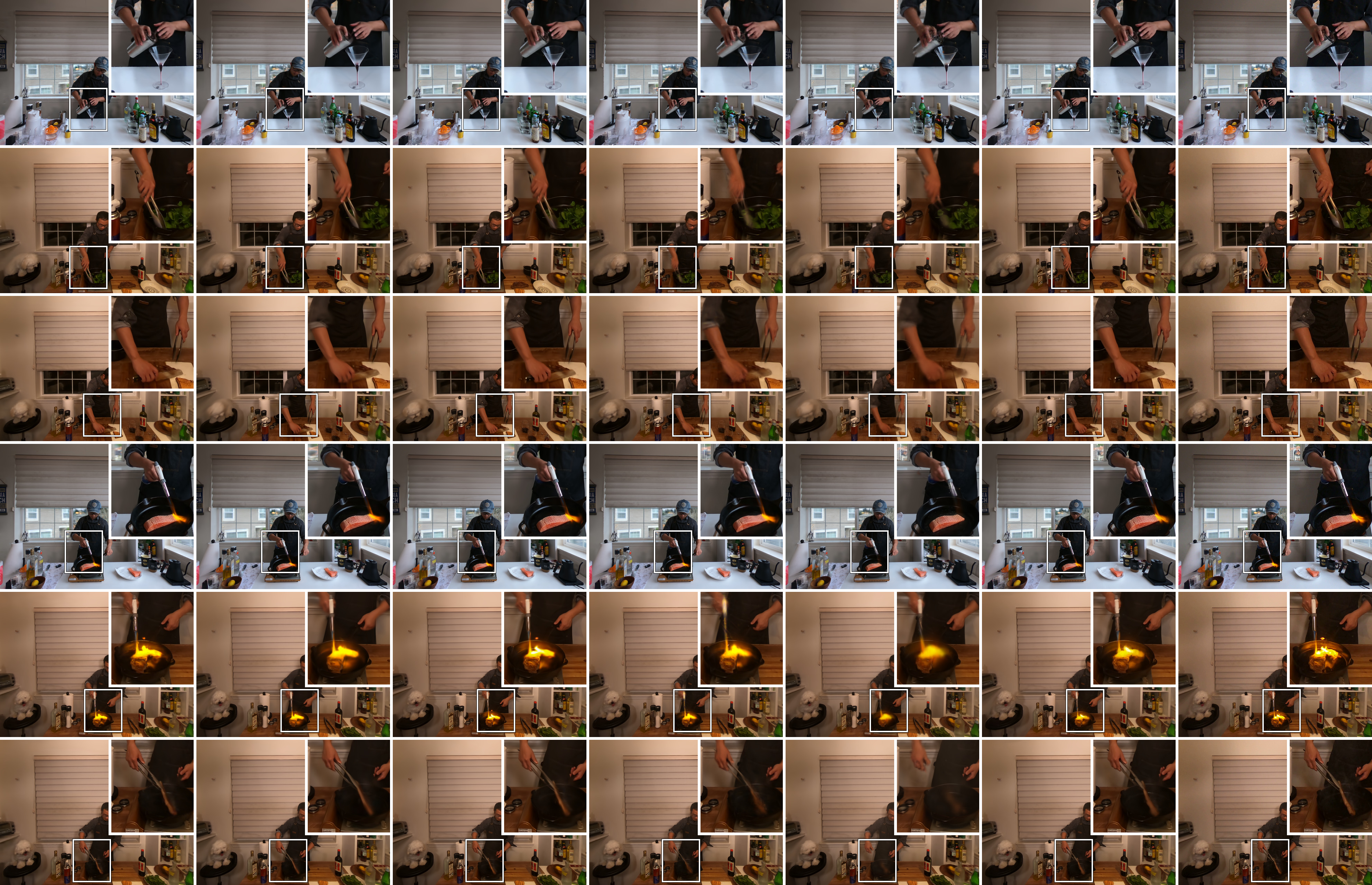}
\caption{%
Qualitative comparison on test views of Neural 3D Video~\cite{li2022neural3dv}. Best viewed digitally and zoomed in.
}\label{fig:qualitative}
\end{figure*}
\begin{figure*}[p]
\centering
\setlength\tabcolsep{0pt}
\footnotesize
\begin{tabular}{*{7}{p{0.14285714\linewidth}<{\centering}}@{}}
4DGS~\cite{yang20244dgaussians} & 4D-Rotor GS~\cite{duan2024rotorgs} & ST Gaussians~\cite{li2024spacetime} & Ex4DGS~\cite{lee2024ex4dgs} & MoRel~\cite{kwak2026morel} & Ours & Ground truth
\end{tabular}
\includegraphics[width=\linewidth]{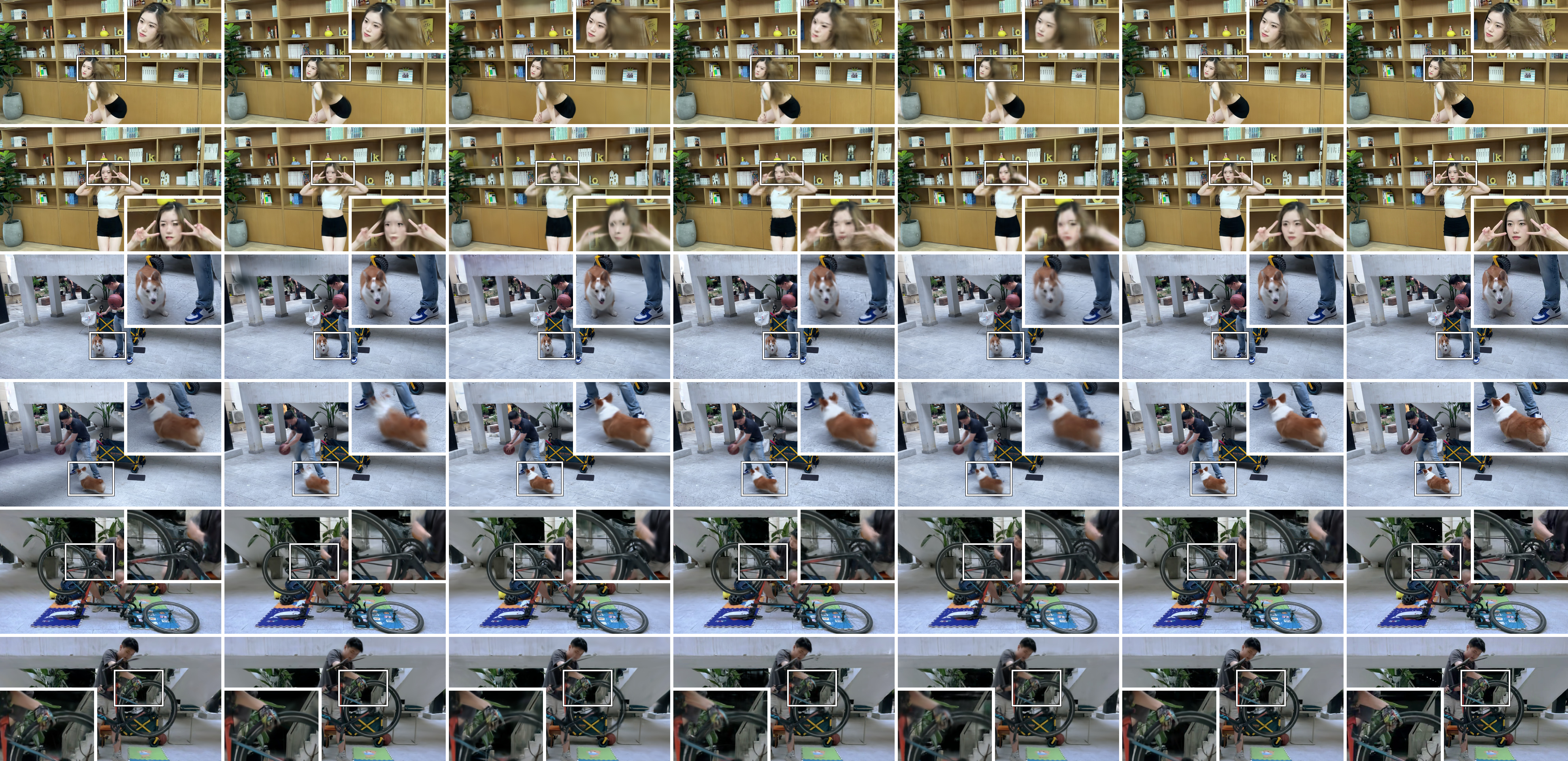}
\caption{%
Qualitative comparison on test views of SelfCap~\cite{wang2025freetimegs}. Best viewed digitally and zoomed in.
}\label{fig:qualitative_selfcap}
\end{figure*}

\begin{table*}[p]
\caption{%
Per-scene results on the Neural 3D Video dataset~\cite{li2022neural3dv} under our protocol, with the methods of \cref{tab:n3dv}. We report mPSNR and mLPIPS, PSNR with the median response of the rig (no calib.), and calibrated PSNR, SSIM, and LPIPS. Time, which includes the initialization as in \cref{tab:n3dv}, and FPS are measured on an RTX 4090. The {\setlength{\fboxsep}{0pt}\colorbox{1st}{\strut\,best\,}} and {\setlength{\fboxsep}{0pt}\colorbox{2nd}{\strut\,second best\,}} results of every scene are color coded.
}\label{tab:per_scene}
\centering
\setlength\tabcolsep{10.73pt}
\renewcommand{\arraystretch}{0.65}
\setlength\extrarowheight{2.5pt}
\setlength\aboverulesep{1.2pt}
\setlength\belowrulesep{2pt}
\scriptsize
\begin{tabular}{llcccccccc}
\toprule
Scene & Method & mPSNR$^\uparrow$ & mLPIPS$^\downarrow$ & No calib.$^\uparrow$ & PSNR$^\uparrow$ & SSIM$^\uparrow$ & LPIPS$^\downarrow$ & Time$^\downarrow$ & FPS$^\uparrow$ \\
\midrule
\multirow{6}{*}{Coffee martini} & 4DGS~\cite{yang20244dgaussians} & \cellcolor{2nd}31.05 & 0.059 & 28.79 & 28.84 & \cellcolor{2nd}0.912 & 0.203 & 2h35m & 81 \\
 & 4D-Rotor GS~\cite{duan2024rotorgs} & 29.25 & 0.090 & 28.42 & 28.52 & 0.909 & 0.207 & 2h43m & 5 \\
 & ST Gaussians~\cite{li2024spacetime} & 30.79 & \cellcolor{2nd}0.053 & 27.39 & 28.80 & \cellcolor{2nd}0.912 & 0.205 & 1h21m & \cellcolor{1st}477 \\
 & Ex4DGS~\cite{lee2024ex4dgs} & 27.47 & 0.137 & 29.16 & 29.20 & 0.911 & 0.203 & \cellcolor{1st}58m & 75 \\
 & MoRel~\cite{kwak2026morel} & 29.15 & 0.083 & \cellcolor{1st}29.44 & \cellcolor{1st}29.73 & \cellcolor{1st}0.923 & \cellcolor{1st}0.182 & 1h40m & 35 \\
 & Ours & \cellcolor{1st}31.92 & \cellcolor{1st}0.039 & \cellcolor{2nd}29.25 & \cellcolor{2nd}29.69 & \cellcolor{1st}0.923 & \cellcolor{2nd}0.183 & \cellcolor{2nd}1h16m & \cellcolor{2nd}255 \\
\midrule
\multirow{6}{*}{Cook spinach} & 4DGS~\cite{yang20244dgaussians} & \cellcolor{1st}31.37 & 0.102 & \cellcolor{2nd}33.15 & 34.25 & 0.954 & 0.198 & 2h19m & 139 \\
 & 4D-Rotor GS~\cite{duan2024rotorgs} & 29.71 & 0.151 & 32.70 & 33.47 & 0.950 & 0.203 & 2h50m & 9 \\
 & ST Gaussians~\cite{li2024spacetime} & 30.85 & \cellcolor{2nd}0.096 & 31.45 & \cellcolor{2nd}34.44 & \cellcolor{2nd}0.956 & 0.195 & 2h55m & \cellcolor{1st}522 \\
 & Ex4DGS~\cite{lee2024ex4dgs} & 29.31 & 0.164 & 32.29 & 33.10 & 0.949 & 0.195 & \cellcolor{1st}57m & 121 \\
 & MoRel~\cite{kwak2026morel} & 28.41 & 0.176 & 32.46 & 33.43 & 0.953 & \cellcolor{2nd}0.190 & 1h32m & 46 \\
 & Ours & \cellcolor{2nd}31.27 & \cellcolor{1st}0.078 & \cellcolor{1st}33.54 & \cellcolor{1st}34.67 & \cellcolor{1st}0.958 & \cellcolor{1st}0.178 & \cellcolor{2nd}1h17m & \cellcolor{2nd}251 \\
\midrule
\multirow{6}{*}{Cut roasted beef} & 4DGS~\cite{yang20244dgaussians} & 34.39 & 0.082 & \cellcolor{2nd}33.26 & \cellcolor{2nd}35.15 & \cellcolor{2nd}0.958 & 0.193 & 2h17m & 136 \\
 & 4D-Rotor GS~\cite{duan2024rotorgs} & 32.23 & 0.130 & 33.11 & 34.16 & 0.953 & 0.199 & 2h47m & 9 \\
 & ST Gaussians~\cite{li2024spacetime} & \cellcolor{2nd}34.45 & \cellcolor{2nd}0.074 & 29.14 & 35.07 & \cellcolor{2nd}0.958 & 0.193 & 2h51m & \cellcolor{1st}540 \\
 & Ex4DGS~\cite{lee2024ex4dgs} & 31.77 & 0.141 & 32.84 & 33.93 & 0.952 & 0.195 & \cellcolor{1st}54m & 131 \\
 & MoRel~\cite{kwak2026morel} & 31.14 & 0.153 & 33.09 & 34.10 & 0.954 & \cellcolor{2nd}0.192 & 1h31m & 46 \\
 & Ours & \cellcolor{1st}35.01 & \cellcolor{1st}0.054 & \cellcolor{1st}33.81 & \cellcolor{1st}35.39 & \cellcolor{1st}0.961 & \cellcolor{1st}0.174 & \cellcolor{2nd}1h16m & \cellcolor{2nd}253 \\
\midrule
\multirow{6}{*}{Flame salmon} & 4DGS~\cite{yang20244dgaussians} & 28.15 & 0.075 & 29.35 & 29.47 & 0.917 & 0.194 & 2h31m & 52 \\
 & 4D-Rotor GS~\cite{duan2024rotorgs} & 27.63 & 0.096 & 28.97 & 29.11 & 0.917 & 0.197 & 2h43m & 6 \\
 & ST Gaussians~\cite{li2024spacetime} & \cellcolor{2nd}28.33 & \cellcolor{2nd}0.070 & 26.96 & 29.77 & 0.923 & 0.191 & 1h40m & \cellcolor{1st}422 \\
 & Ex4DGS~\cite{lee2024ex4dgs} & 26.85 & 0.128 & 29.23 & 29.40 & 0.916 & 0.190 & \cellcolor{1st}56m & 75 \\
 & MoRel~\cite{kwak2026morel} & 26.69 & 0.093 & \cellcolor{2nd}29.89 & \cellcolor{2nd}30.29 & \cellcolor{2nd}0.928 & \cellcolor{2nd}0.172 & 1h41m & 35 \\
 & Ours & \cellcolor{1st}29.09 & \cellcolor{1st}0.054 & \cellcolor{1st}30.26 & \cellcolor{1st}30.44 & \cellcolor{1st}0.930 & \cellcolor{1st}0.170 & \cellcolor{2nd}1h14m & \cellcolor{2nd}257 \\
\midrule
\multirow{6}{*}{Flame steak} & 4DGS~\cite{yang20244dgaussians} & 32.03 & 0.067 & \cellcolor{2nd}34.24 & 35.23 & \cellcolor{2nd}0.961 & 0.183 & 2h20m & 143 \\
 & 4D-Rotor GS~\cite{duan2024rotorgs} & 31.07 & 0.097 & 33.79 & 34.57 & 0.958 & 0.190 & 2h46m & 9 \\
 & ST Gaussians~\cite{li2024spacetime} & \cellcolor{1st}32.24 & \cellcolor{2nd}0.058 & 29.69 & \cellcolor{1st}35.92 & \cellcolor{1st}0.964 & 0.182 & 2h53m & \cellcolor{1st}537 \\
 & Ex4DGS~\cite{lee2024ex4dgs} & 30.10 & 0.110 & 32.88 & 34.27 & 0.957 & 0.184 & \cellcolor{1st}58m & 147 \\
 & MoRel~\cite{kwak2026morel} & 29.75 & 0.107 & 33.43 & 34.66 & \cellcolor{2nd}0.961 & \cellcolor{2nd}0.177 & 1h33m & 45 \\
 & Ours & \cellcolor{2nd}32.20 & \cellcolor{1st}0.048 & \cellcolor{1st}35.04 & \cellcolor{2nd}35.70 & \cellcolor{1st}0.964 & \cellcolor{1st}0.170 & \cellcolor{2nd}1h17m & \cellcolor{2nd}253 \\
\midrule
\multirow{6}{*}{Sear steak} & 4DGS~\cite{yang20244dgaussians} & 35.38 & 0.064 & \cellcolor{2nd}33.99 & 35.75 & \cellcolor{2nd}0.962 & 0.186 & 2h18m & 136 \\
 & 4D-Rotor GS~\cite{duan2024rotorgs} & 33.92 & 0.091 & 33.66 & 35.09 & 0.959 & 0.191 & 2h47m & 9 \\
 & ST Gaussians~\cite{li2024spacetime} & \cellcolor{2nd}35.80 & \cellcolor{2nd}0.054 & 30.44 & \cellcolor{1st}36.61 & \cellcolor{1st}0.965 & \cellcolor{2nd}0.183 & 2h54m & \cellcolor{1st}545 \\
 & Ex4DGS~\cite{lee2024ex4dgs} & 34.12 & 0.087 & 32.41 & 34.47 & 0.956 & 0.186 & \cellcolor{1st}55m & 147 \\
 & MoRel~\cite{kwak2026morel} & 32.37 & 0.102 & 32.64 & 34.34 & 0.956 & 0.190 & 1h29m & 55 \\
 & Ours & \cellcolor{1st}36.16 & \cellcolor{1st}0.043 & \cellcolor{1st}34.98 & \cellcolor{2nd}36.33 & \cellcolor{1st}0.965 & \cellcolor{1st}0.173 & \cellcolor{2nd}1h18m & \cellcolor{2nd}243 \\
\bottomrule
\end{tabular}
\end{table*}

\begin{table*}[p]
\caption{%
Per-scene results on the SelfCap dataset~\cite{wang2025freetimegs} under our protocol, with the columns and color coding of \cref{tab:per_scene}.
}\label{tab:per_scene_selfcap}
\centering
\setlength\tabcolsep{10.73pt}
\renewcommand{\arraystretch}{0.65}
\setlength\extrarowheight{2.5pt}
\setlength\aboverulesep{1.2pt}
\setlength\belowrulesep{2pt}
\scriptsize
\begin{tabular}{llcccccccc}
\toprule
\rlap{Scene}\phantom{Cut roasted beef} & Method & mPSNR$^\uparrow$ & mLPIPS$^\downarrow$ & No calib.$^\uparrow$ & PSNR$^\uparrow$ & SSIM$^\uparrow$ & LPIPS$^\downarrow$ & Time$^\downarrow$ & FPS$^\uparrow$ \\
\midrule
\multirow{6}{*}{Dance 1} & 4DGS~\cite{yang20244dgaussians} & \cellcolor{2nd}26.44 & \cellcolor{2nd}0.191 & 22.68 & \cellcolor{2nd}27.50 & 0.912 & 0.198 & 2h59m & 103 \\
 & 4D-Rotor GS~\cite{duan2024rotorgs} & 25.46 & 0.236 & \cellcolor{2nd}23.35 & \cellcolor{2nd}27.50 & \cellcolor{1st}0.919 & \cellcolor{1st}0.174 & 6h40m & 3 \\
 & ST Gaussians~\cite{li2024spacetime} & 23.71 & 0.271 & 20.51 & 25.53 & 0.896 & 0.254 & \cellcolor{2nd}40m & \cellcolor{1st}354 \\
 & Ex4DGS~\cite{lee2024ex4dgs} & 24.12 & 0.275 & 22.57 & 26.97 & 0.909 & 0.185 & 1h16m & 96 \\
 & MoRel~\cite{kwak2026morel} & 22.62 & 0.329 & \cellcolor{1st}23.77 & 26.16 & \cellcolor{2nd}0.916 & 0.186 & 2h50m & 39 \\
 & Ours & \cellcolor{1st}26.73 & \cellcolor{1st}0.167 & 22.12 & \cellcolor{1st}27.76 & \cellcolor{1st}0.919 & \cellcolor{2nd}0.176 & \cellcolor{1st}34m & \cellcolor{2nd}205 \\
\midrule
\multirow{6}{*}{Dance 2} & 4DGS~\cite{yang20244dgaussians} & 23.54 & \cellcolor{2nd}0.225 & 22.42 & 26.96 & 0.908 & 0.206 & 2h58m & 105 \\
 & 4D-Rotor GS~\cite{duan2024rotorgs} & \cellcolor{2nd}23.96 & 0.251 & \cellcolor{2nd}22.51 & \cellcolor{1st}27.00 & \cellcolor{1st}0.915 & \cellcolor{1st}0.179 & 6h22m & 3 \\
 & ST Gaussians~\cite{li2024spacetime} & 22.04 & 0.329 & 20.55 & 25.50 & 0.895 & 0.248 & \cellcolor{2nd}40m & \cellcolor{1st}368 \\
 & Ex4DGS~\cite{lee2024ex4dgs} & 21.36 & 0.324 & 22.06 & 26.20 & 0.904 & 0.188 & 1h14m & 101 \\
 & MoRel~\cite{kwak2026morel} & 20.60 & 0.356 & \cellcolor{1st}23.05 & 25.68 & 0.905 & 0.203 & 2h46m & 36 \\
 & Ours & \cellcolor{1st}24.46 & \cellcolor{1st}0.202 & 21.75 & \cellcolor{2nd}26.98 & \cellcolor{2nd}0.913 & \cellcolor{2nd}0.180 & \cellcolor{1st}33m & \cellcolor{2nd}207 \\
\midrule
\multirow{6}{*}{Corgi 1} & 4DGS~\cite{yang20244dgaussians} & \cellcolor{2nd}25.05 & \cellcolor{2nd}0.115 & \cellcolor{2nd}26.46 & 27.07 & 0.853 & 0.201 & 2h57m & 94 \\
 & 4D-Rotor GS~\cite{duan2024rotorgs} & 24.66 & 0.143 & 21.83 & 22.28 & \cellcolor{1st}0.886 & \cellcolor{1st}0.187 & 9h15m & 2 \\
 & ST Gaussians~\cite{li2024spacetime} & 24.74 & 0.158 & 24.48 & 26.50 & 0.800 & 0.336 & \cellcolor{2nd}37m & \cellcolor{1st}481 \\
 & Ex4DGS~\cite{lee2024ex4dgs} & 22.51 & 0.301 & 25.08 & 26.07 & 0.789 & 0.300 & 1h03m & 203 \\
 & MoRel~\cite{kwak2026morel} & 23.46 & 0.234 & 26.02 & \cellcolor{2nd}27.89 & 0.856 & 0.249 & 2h39m & 49 \\
 & Ours & \cellcolor{1st}25.90 & \cellcolor{1st}0.112 & \cellcolor{1st}26.73 & \cellcolor{1st}27.98 & \cellcolor{2nd}0.885 & \cellcolor{2nd}0.192 & \cellcolor{1st}33m & \cellcolor{2nd}206 \\
\midrule
\multirow{6}{*}{Corgi 2} & 4DGS~\cite{yang20244dgaussians} & 20.85 & \cellcolor{2nd}0.247 & 19.09 & 20.69 & 0.842 & 0.217 & 2h58m & 94 \\
 & 4D-Rotor GS~\cite{duan2024rotorgs} & 20.87 & 0.313 & \cellcolor{1st}27.15 & \cellcolor{1st}27.87 & \cellcolor{1st}0.880 & \cellcolor{1st}0.194 & 8h10m & 2 \\
 & ST Gaussians~\cite{li2024spacetime} & \cellcolor{2nd}21.84 & 0.273 & 24.62 & 26.66 & 0.798 & 0.341 & \cellcolor{2nd}39m & \cellcolor{1st}469 \\
 & Ex4DGS~\cite{lee2024ex4dgs} & 21.30 & 0.342 & 26.05 & 26.72 & 0.800 & 0.293 & 1h04m & 194 \\
 & MoRel~\cite{kwak2026morel} & 20.00 & 0.370 & 25.55 & 26.44 & 0.848 & 0.260 & 2h36m & 50 \\
 & Ours & \cellcolor{1st}22.17 & \cellcolor{1st}0.244 & \cellcolor{2nd}26.85 & \cellcolor{2nd}27.74 & \cellcolor{2nd}0.876 & \cellcolor{2nd}0.203 & \cellcolor{1st}33m & \cellcolor{2nd}206 \\
\midrule
\multirow{6}{*}{Bike 1} & 4DGS~\cite{yang20244dgaussians} & 25.70 & 0.095 & 26.22 & 27.07 & 0.869 & 0.249 & 1h42m & 175 \\
 & 4D-Rotor GS~\cite{duan2024rotorgs} & \cellcolor{2nd}25.87 & \cellcolor{2nd}0.093 & \cellcolor{2nd}26.85 & 27.87 & \cellcolor{1st}0.893 & \cellcolor{1st}0.207 & 2h30m & 9 \\
 & ST Gaussians~\cite{li2024spacetime} & 25.67 & 0.104 & 26.43 & \cellcolor{2nd}27.91 & 0.888 & 0.248 & \cellcolor{2nd}25m & \cellcolor{1st}674 \\
 & Ex4DGS~\cite{lee2024ex4dgs} & 25.02 & 0.128 & 26.55 & 27.43 & 0.865 & 0.256 & 44m & 253 \\
 & MoRel~\cite{kwak2026morel} & 24.37 & 0.143 & 26.84 & 27.88 & 0.881 & 0.235 & 1h47m & 57 \\
 & Ours & \cellcolor{1st}26.09 & \cellcolor{1st}0.080 & \cellcolor{1st}27.27 & \cellcolor{1st}27.99 & \cellcolor{2nd}0.890 & \cellcolor{2nd}0.208 & \cellcolor{1st}22m & \cellcolor{2nd}271 \\
\midrule
\multirow{6}{*}{Bike 2} & 4DGS~\cite{yang20244dgaussians} & 25.22 & 0.125 & 25.97 & 27.40 & 0.882 & 0.241 & 1h47m & 159 \\
 & 4D-Rotor GS~\cite{duan2024rotorgs} & \cellcolor{2nd}25.87 & \cellcolor{2nd}0.121 & \cellcolor{2nd}26.44 & \cellcolor{1st}28.53 & \cellcolor{1st}0.903 & \cellcolor{2nd}0.209 & 1h51m & 13 \\
 & ST Gaussians~\cite{li2024spacetime} & 23.02 & 0.210 & 25.41 & 26.75 & 0.871 & 0.272 & \cellcolor{2nd}26m & \cellcolor{1st}665 \\
 & Ex4DGS~\cite{lee2024ex4dgs} & 24.51 & 0.175 & 25.79 & 27.45 & 0.870 & 0.259 & 44m & 269 \\
 & MoRel~\cite{kwak2026morel} & 24.03 & 0.204 & 25.78 & 27.56 & 0.890 & 0.219 & 1h43m & 59 \\
 & Ours & \cellcolor{1st}25.99 & \cellcolor{1st}0.104 & \cellcolor{1st}26.66 & \cellcolor{2nd}28.15 & \cellcolor{2nd}0.900 & \cellcolor{1st}0.204 & \cellcolor{1st}22m & \cellcolor{2nd}284 \\
\bottomrule
\end{tabular}
\end{table*}

\subsection{Qualitative Results}
\label{sec:suppl_qualitative}
\Cref{fig:qualitative,fig:qualitative_selfcap} compare all methods of \cref{tab:n3dv} on test views of Neural 3D Video and SelfCap.
The methods differ most on fast-moving content, such as hands, hair, and the dog.

\subsection{Per-Scene Results}
\label{sec:suppl_per_scene}
\Cref{tab:per_scene,tab:per_scene_selfcap} list the per-scene numbers behind \cref{tab:n3dv}.

\end{document}